\documentclass[11pt]{idea_style}
\newif\ifdraft
\draftfalse

\usepackage[T1]{fontenc}
\usepackage{url}
\usepackage{nicefrac}
\usepackage{multicol}
\usepackage{microtype}
\usepackage{geometry}
\usepackage{graphicx}%
\graphicspath{{figures/}{figures/exp/}}
\usepackage{multirow}%
\usepackage{amsmath}
\usepackage{amssymb}
\usepackage{amsfonts}
\usepackage{stmaryrd}
\usepackage{mathrsfs}%
\usepackage{xcolor}%
\usepackage{textcomp}%
\usepackage{manyfoot}%
\usepackage{booktabs}%
\usepackage{algorithm}%
\usepackage{tabularx}%
\usepackage{algorithmicx}%
\usepackage{algpseudocode}%
\usepackage{listings}%
\usepackage[inline]{enumitem}
\usepackage{xspace}
\usepackage{tikz}
\usepackage{makecell}
\usepackage{longtable}
\usepackage{changepage}
\usepackage{color}
\usepackage{colortbl}
\usepackage{pifont}
\usepackage{subcaption}
\usepackage{physics}
\usepackage{diagbox}
\usepackage{wrapfig}
\usepackage{siunitx}
\usepackage[misc]{ifsym}
\usepackage{threeparttable}
\usepackage{cutwin}
\usepackage{dsfont}
\usepackage{float}
\usepackage{tablefootnote}
\usepackage{placeins}
\usepackage[numbers]{natbib}

\usepackage{hyperref}
\hypersetup{
  colorlinks,
  linkcolor=ideacolor,
  citecolor=ideacolor,
  urlcolor=ideacolor
}
\usepackage[capitalize,nameinlink]{cleveref}

\usepackage{setspace}
\usepackage{changepage}

\usepackage{amsmath,amsfonts,bm}

\def\eqref#1{equation~\ref{#1}}
\def\1{\bm{1}}

\DeclareMathAlphabet{\mathsfit}{\encodingdefault}{\sfdefault}{m}{sl}
\SetMathAlphabet{\mathsfit}{bold}{\encodingdefault}{\sfdefault}{bx}{n}

\definecolor{myorange}{RGB}{220,128,0}
\definecolor{myblue}{RGB}{0,80,220}
\definecolor{myred}{RGB}{220,20,0}
\definecolor{deepred}{RGB}{180,40,40}
\definecolor{deepblue}{RGB}{40,90,200}
\definecolor{deeppurple}{RGB}{75,0,130}

\newcommand{\circledmark}[2]{%
  \tikz[baseline=(char.base)]{%
    \node[
      shape=circle,
      fill=#1,
      text=white,
      inner sep=0.6pt,
      minimum size=1.55ex,
      font=\scriptsize\bfseries
    ] (char) {#2};%
  }%
}

\def\ourname{GRAPE}
\newcommand{\methodname}{\ourname}

\newcommand{\dsLsThreeDw}{LS3DW}
\newcommand{\dsCelebA}{CelebA}
\newcommand{\dsLaPa}{LaPa}
\newcommand{\dsLfw}{LFW}
\newcommand{\dsHdtf}{HDTF}
\newcommand{\dsNer}{NersembleV2}

\newcommand{\mdown}{$\downarrow$}
\newcommand{\mup}{$\uparrow$}
\newcommand{\best}[1]{\textbf{#1}}
\newcommand{\second}[1]{\underline{#1}}

\ifdraft
    \providecommand\todo[1]{[\textcolor{red}{TODO: {#1}}]}
\else
    \providecommand\todo[1]{}
\fi

\author[]{Yunfei Liu$^{(\textrm{\Letter})}$, Lijian Lin, Ye Zhu, Yu Li}

\affiliation{International Digital Economy Academy}

\ideadata[$^{(\textrm{\Letter})}$ ]{Corresponding author.}
\ideadata[Keywords]{Portrait mesh estimation, Talking head, Anatomical disentanglement}

\title{GRAPE: Graduated Routing for Articulated Portrait mesh Estimation}

\abstract{
\small{
Articulated portrait mesh estimation is fundamental to 3D understanding, avatar generation, and immersive interaction. Existing approaches primarily rely on 3D Morphable Models (3DMMs). However, face-centric models suffer from the ``floating head'' assumption, conflating head pose with global rotation due to the lack of neck kinematics. Conversely, body-centric models lack high-fidelity facial expression capabilities. Furthermore, current methods struggle to disentangle jaw articulation from expression blendshapes, often over-relying on expressions for mouth opening. These limitations make monocular portrait recovery difficult across representation, supervision, and anatomical parameter estimation. To address these limitations, we introduce \textbf{GRAPE} (Graduated Routing for Articulated Portrait mesh Estimation). We build a \textbf{Portrait Parametric Model (PPM)} with an explicit torso-to-head kinematic chain and a canonical injection step to merge FLAME and the SMPL-X torso. We propose a \textbf{Progressive Anatomical Alignment (PAA)} network, which is composed of a pretrained portrait encoder, a Graduated-Mask Router, and coarse-to-fine experts that follow the portrait anatomical prior. We then train this network with multi-source supervision that combines sparse anatomical keypoints, feature distillation, foreground mask constraints, and relative geometry constraints.
Experiments show that GRAPE improves portrait mesh recovery quality, pose alignment, and jaw--expression disentanglement over prior methods. We also demonstrate that our method can benefit the downstream tasks of audio-driven talking-head generation and 3D portrait generation.

}
}

\begin{document}

\maketitle

\begin{figure}[h]
    \centering
    \includegraphics[width=\textwidth]{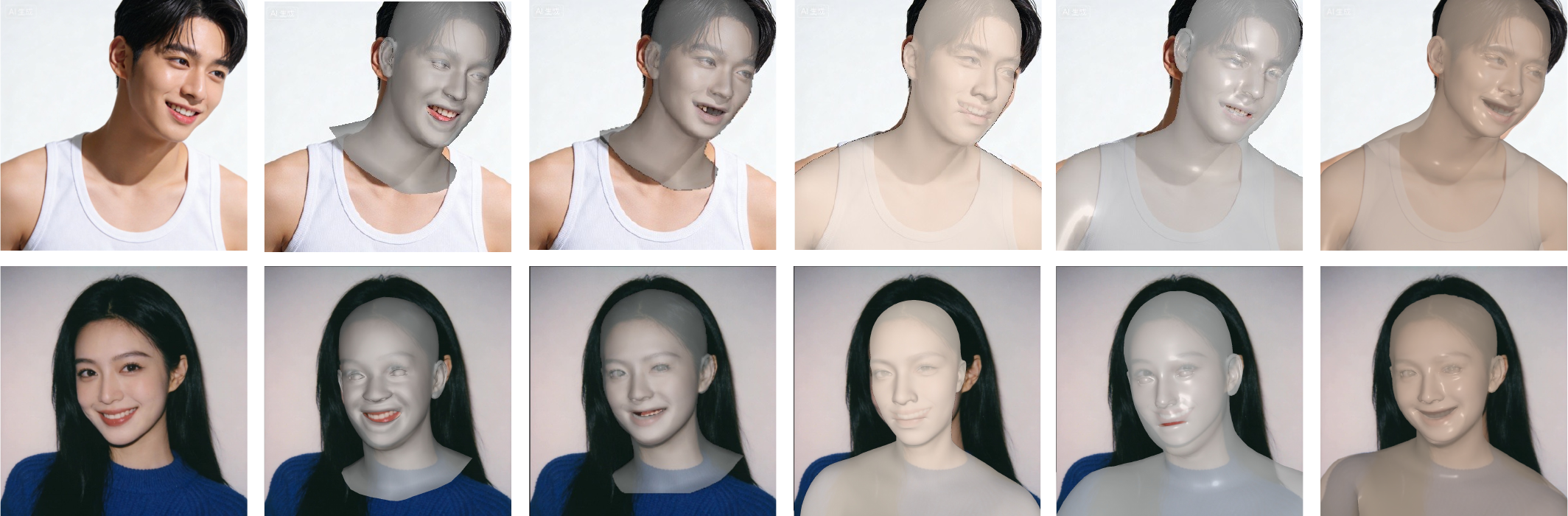}\\[-0.4em]
    {\footnotesize
    \noindent
    \begin{tabular}{@{}*{6}{>{\centering\arraybackslash}p{\dimexpr\textwidth/6\relax}@{}}}
      Input Image
      & SPECTRE~\cite{filntisis2022visual}
      & TEASER~\cite{Liu2025TEASER}
      & SAM-3DB~\cite{yang2026sam3dbody}
      & SMPLest-X~\cite{yin2025smplest}
      & Ours \\
    \end{tabular}}
    \caption{We propose \textbf{GRAPE} (Graduated Routing for Articulated Portrait mesh Estimation). The input portrait images are AI generated.}
    \label{fig:teaser}
\end{figure}

% !TEX root = ../main.tex
\section{Introduction}
\label{sec:intro}

Articulated portrait mesh estimation is pivotal for 3D scene understanding, immersive telepresence, and the animation of digital avatars. While 3D Morphable Models (3DMMs) have become the standard for representing facial geometry, recovering high-fidelity, animatable portraits from monocular images remains an ill-posed problem. Current approaches predominantly rely on face-centric models (e.g., FLAME \cite{li2017learning}) or full-body models (e.g., SMPL-X \cite{pavlakos2019expressive}). However, despite significant advances in shape reconstruction accuracy, existing methods struggle to achieve \textit{kinematic disentanglement}---the ability to independently and accurately control head pose, jaw articulation, and surface expression. {We view this difficulty as a chain of three under-specifications in current portrait recovery pipelines: a representation gap, a supervision gap, and a parameter-factorization gap.}

{The first gap is representational.}
Most state-of-the-art face reconstruction methods \cite{deng2019accurate,feng2021learning,zielonka2022towards} crop the input to the facial region, discarding the neck and torso. {Even when their camera parameterization differs, their reconstruction is effectively head-centered because the model lacks a torso-rooted kinematic reference.} Without the torso as a kinematic anchor, the decomposition of global rotation into intrinsic neck articulation and extrinsic camera rotation becomes mathematically ambiguous. This often results in ``centrifuge-like'' head rotations in synthesized animations, where the head rotates around its center rather than the anatomical neck joint.
These kinematic inconsistencies also make existing talking-avatar models harder to train and generalize, as motion must be learned without a stable head--torso reference frame.

{The second gap appears after introducing a more complete portrait representation: the torso, neck, and shoulder variables are much more weakly supervised than the face in in-the-wild images.} Sparse keypoints from human foundation models such as Sapiens-2~\cite{sapiens2} provide useful face and shoulder cues, but they can be noisy or missing under cropping, clothing, occlusion, and profile views. {Therefore, a portrait model should not rely on a single pseudo-label source; it needs complementary supervision in feature space, projection space, and relative 3D geometry.}

{The third gap is parameter factorization.} In the FLAME parameter space, mouth opening can be approximated either by rotating the jaw joint or by activating expression coefficients. Since neural networks prioritize shortcut learning, regressors often over-rely on expression parameters to fit mouth movements. This leads to ``unanimatable'' meshes where the jaw bone remains static during speech, violating human physiology. While recent works attempt to mitigate this via soft regularization (e.g., lip-reading losses \cite{filntisis2022visual} or emotion consistency \cite{danve2022emoca}), they fail to impose strict structural constraints.
To make this failure mode measurable on talking-head clips with relatively stable head pose, {we use the normalized projected nose-tip-to-chin-tip distance as a weak skeletal reference for jaw opening, since it is less directly controlled by local lip-expression blendshapes than mouth-landmark distance.}
During talking, jaw pitch should co-vary with this distance; yet for recent regressors, predicted jaw rotation remains weakly correlated while expression coefficients absorb most mouth motion (\cref{fig:jaw_distance_analysis}).
This mismatch explains why meshes can look plausible in a single frame but fail under jaw-driven re-animation.

To address these limitations, {we formulate portrait recovery as a progressive resolution of the three gaps above rather than as a flat parameter regression problem.} We propose \textbf{graduated routing}: features and supervision follow the biological hierarchy from global pose to local expression. Human motion adheres to a strict ordering: the torso anchors the neck, the neck drives the head, and the jaw articulates before soft tissues deform. Ignoring this \textit{hierarchical causality} leads to the aforementioned coupling artifacts.

Based on this insight, we present \textbf{GRAPE} (Graduated Routing for Articulated Portrait mesh Estimation). First, we introduce the \textbf{Portrait Parametric Model (PPM)}, which attaches a FLAME head to an SMPL-X torso and defines pose from the spine to the head. Second, we propose \textbf{Progressive Anatomical Alignment (PAA)} via the HKD-Exp Network: a frozen Sapiens-2 encoder~\cite{sapiens2} extracts features, three experts regress PPM parameters from coarse to fine, and a \textbf{Graduated-Mask Router} randomly drops deeper supervision during training to reduce shortcut learning. {Third, to supervise the richer but weakly labeled portrait variables, we use multi-source supervision: Sapiens-2 keypoints for sparse anatomical anchors, Pixel3DMM feature distillation for representation-level geometry priors, foreground projection masks for visible portrait extent, and relative geometry losses for head--torso layout.} 
% Finally, we train on pseudo-labels generated by our offline \textbf{Graduated-Refinement Pipeline}.

In summary, our contributions are:
\begin{itemize}
    \item We define the Portrait Parametric Model (PPM), a unified head-neck-shoulder representation that resolves the pose ambiguity inherent in floating-head approaches.
    \item We propose Progressive Anatomical Alignment (PAA) with a Graduated-Mask Router to regress pose before expression and reduce jaw--expression entanglement.
    \item {We introduce multi-source supervision for PPM, combining Sapiens-2 keypoints, Pixel3DMM feature distillation, foreground projection masks, and relative geometry constraints to stabilize weakly supervised portrait variables.}
    \item Experiments show that our approach outperforms existing methods on multiple datasets and benefits downstream audio-driven talking avatars and 3D portrait reconstruction.
\end{itemize}

% !TEX root = ../main.tex
\section{Related Work}
\label{sec:related}

\textbf{Monocular Face Reconstruction and The ``Floating Head'' Ambiguity.}
Recent state-of-the-art methods predominantly frame reconstruction as a regression task using 3D Morphable Models (3DMMs)~\cite{blanz1999morphable}. DECA~\cite{feng2021learning} utilizes detailed displacement maps to capture high-frequency surface details, while MICA~\cite{zielonka2022towards} focuses on metric-level shape recovery by leveraging recognition networks. Recent regressors such as TEASER~\cite{Liu2025TEASER} further improve expression fidelity with token-based spatial modeling, but still operate on tightly cropped faces.
{GNM~\cite{google2026gnm} provides an expressive head model with internal anatomy such as teeth and tongue.}
{However, its jaw motion remains entangled with expression, and the camera is still centered at the head.}
These face-centric approaches~\cite{deng2019accurate,feng2021learning,zielonka2022towards,ren2021pirender,Liu2025TEASER,google2026gnm} discard the neck and torso. As noted in~\cite{pavlakos2019expressive,feng2021collaborative}, this ``floating head'' assumption renders the decomposition of global rotation into intrinsic neck articulation and extrinsic camera pose mathematically ill-posed. Without a stable head--torso reference frame, downstream talking-head and avatar models~\cite{cudeiro2019capture,wang2020mead} are also harder to train and generalize. While full-body methods like SMPL-X~\cite{pavlakos2019expressive}, {PIXIE}~\cite{feng2021collaborative}, and {OSX}~\cite{lin2023one} explicitly model the kinematic chain from the spine, they often compromise facial fidelity. The facial topology in SMPL-X is a lower-dimensional approximation of FLAME, lacking the expressive blendshapes required for nuanced talking avatars. Our PPM topology bridges this gap, integrating the kinematic stability of SMPL-X with the high-fidelity geometry of FLAME~\cite{li2017learning}.

\textbf{Disentangled Animation and Expression Control.}
A core challenge in animatable reconstruction is disentangling identity, pose, and expression. {EMOCA}~\cite{danve2022emoca} significantly improves emotional fidelity by supervising expression regression with a deep emotion recognition loss. Similarly, {SPECTRE}~\cite{filntisis2022visual} enforces audio-visual consistency via a lip-reading network to capture accurate mouth articulations. {TEASER}~\cite{Liu2025TEASER} achieves strong landmark and expression fitting, yet its predicted jaw pose can remain weakly correlated with skeletal motion during speech when expression blendshapes absorb mouth opening. Despite these semantic regularizations, prior methods lack structural constraints. In FLAME-based optimization~\cite{li2017learning} and regression~\cite{danve2022emoca}, jaw rotation ($\boldsymbol{\theta}_{jaw}$) and mouth-opening blendshapes ($\boldsymbol{\psi}$) are often coupled. Neural regressors frequently converge to a local optimum where expressions compensate for static jaw poses, resulting in ``dead jaw'' artifacts during re-animation~\cite{filntisis2022visual,danve2022emoca}. Moreover, appearance-based metrics such as landmark mouth opening can be inflated by expression shortcuts and therefore mis-rank kinematic disentanglement. Unlike these soft regularization approaches, our \textbf{Graduated-Mask Router} imposes a hard architectural constraint, forcing the network to prioritize rigid articulation before refining non-rigid deformations.

\textbf{Pseudo-Labeling for Weakly-Supervised Learning.}
Due to the scarcity of in-the-wild 3D ground truth, self-supervised learning via differentiable rendering is standard practice~\cite{tewari2017mofa,deng2019accurate,sanyal2019learning}. However, photometric losses are prone to depth-scale ambiguities and often fail to separate texture from lighting. Recent trends involve generating pseudo-labels via optimization-based fitting~\cite{feng2021learning,lin2023one,qiu2022sculptor}, often warm-started from strong regressors such as TEASER~\cite{Liu2025TEASER} for the face and ProHMR~\cite{Kolotouros2021ProHMR} for the body. Yet, naive joint fitting often inherits the entanglement of the underlying model (e.g., fitting a smile with jaw rotation). Our \textbf{Graduated-Refinement Pipeline} differs by strictly staging the optimization process (Global $\rightarrow$ Articulation $\rightarrow$ Details) to generate structurally decoupled pseudo-labels, providing cleaner supervision for the regressor.

\textbf{Foundation Priors and Feature Distillation.}
Large-scale pretrained human models have recently been adopted as priors for monocular reconstruction. {Pixel3DMM}~\cite{giebenhain2025pixel3dmm} distills screen-space geometric cues into regressors for single-image 3D face recovery, while {Sapiens-2}~\cite{sapiens2} provides dense human keypoints from large-scale pretraining. Most prior work applies such features through flat regression heads over a single 3DMM parameter vector. Our method instead routes frozen Sapiens-2 features and Pixel3DMM distillation through \textbf{Progressive Anatomical Alignment}, attaching representation-level priors to the anatomically ordered experts of PPM rather than mixing shape, articulation, and expression in one prediction stage.

% !TEX root = ../main.tex
\section{Method}
\label{sec:method}

Given a portrait image $\mathbf{I}$, we estimate decoupled parameters of the Portrait Parametric Model (PPM) and reconstruct an animatable portrait mesh. Specifically, our method addresses three linked gaps. PPM closes the representation gap by replacing a head-centered face model with a torso-rooted portrait model. Progressive Anatomical Alignment (PAA) closes the factorization gap by routing parameters in the same coarse-to-fine order as the portrait kinematic tree. Multi-source and ordinal supervision closes the supervision gap by combining sparse anatomical keypoints with feature-, mask-, and geometry-level constraints. \cref{subsec:ppm} defines PPM; \cref{subsec:paa} describes PAA; \cref{subsec:impl} summarizes data and training.

\subsection{Portrait Parametric Model}
\label{subsec:ppm}

Most 3DMM face models treat head pose as a single global rotation. {This is sufficient for cropped face alignment but under-specifies a portrait, where the visible neck and shoulders define the natural reference frame for head motion.} Physically, head pose is the rotation of the neck relative to the torso, while camera pose is an independent extrinsic variable. Without a torso anchor, these quantities cannot be reliably separated, which leads to the ``floating head'' problem in monocular reconstruction and animation.

Full-body models such as SMPL-X provide a torso anchor, but their built-in face space is limited: the body shape space has only 10 dimensions, and the facial expression space is too low-dimensional to capture fine talking-face motion. We therefore build a hybrid model $\mathcal{M}_{\mathrm{PPM}}$ that keeps the FLAME2023 head for shape and expression~\cite{li2017learning,flame23}, and attaches it to the SMPL-X~\cite{pavlakos2019expressive} torso (SMPL-T) for neck and shoulder kinematics. Specifically, we remove the lower-body and arm meshes and update the SMPL-X LBS weights to simplify the torso representation.

\begin{figure*}[t]
    \centering
    \includegraphics[width=\linewidth]{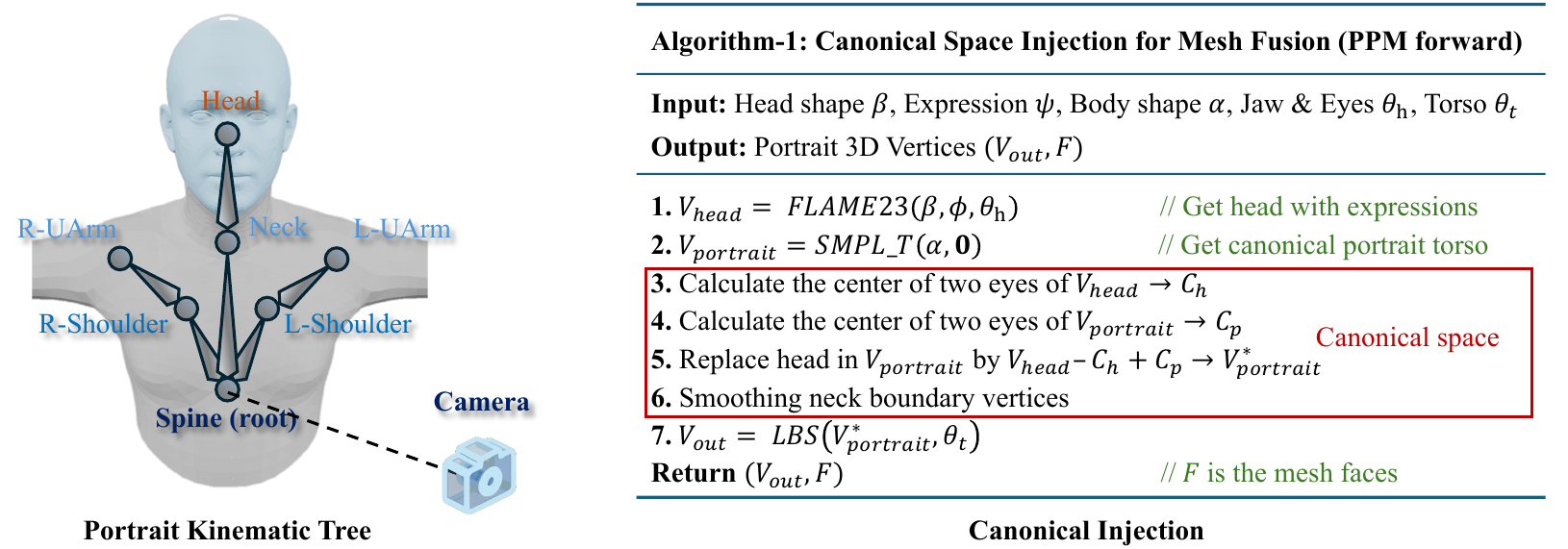}
    \caption{\textbf{Portrait Parametric Model (PPM).} \textbf{Left:} Kinematic tree rooted at the spine; the torso anchors camera-relative pose. \textbf{Right:} Canonical Injection. The FLAME head and SMPL-T are aligned in canonical space via eye centers ($C_h \rightarrow C_p$) and merged into one watertight mesh.}
    \label{fig:ppm_pipeline}
\end{figure*}

\subsubsection{Kinematic-aware Topology.}
PPM maps portrait parameters to a mesh $\mathbf{P}=\{\mathbf{V}, \mathbf{F}\}$:
\begin{equation}
    \mathbf{P} = \mathrm{PPM}(\boldsymbol{\beta}_h, \boldsymbol{\beta}_t, \mathbf{s}, \boldsymbol{\psi}, \boldsymbol{\theta}),
\end{equation}
where $\mathbf{V} \in \mathbb{R}^{N_v \times 3}$ and $\mathbf{F} \in \mathbb{N}^{N_f \times 3}$ are vertices and faces. Head shape $\boldsymbol{\beta}_h$ and expression $\boldsymbol{\psi}$ come from FLAME2023~\cite{flame23}; torso shape $\boldsymbol{\beta}_t$ comes from SMPL-X~\cite{pavlakos2019expressive}; $\mathbf{s}$ is a global scale of the head mesh; pose $\boldsymbol{\theta}$ covers the spine (root), neck, head, jaw, and eyes. We also add shoulder and upper-arm poses for torso movements. As shown in \cref{fig:ppm_pipeline} (Left), the forward kinematic tree for the head is
\begin{equation}
    \text{Spine (Root)} \rightarrow \text{Neck} \rightarrow \text{Head} \rightarrow \{\text{Jaw, Eyes}\}.
\end{equation}
The camera is defined relative to the spine root rather than the head center. Because the head joint is a child of the neck ($\mathbf{J}_{\mathrm{head}} \in \mathrm{child}(\mathbf{J}_{\mathrm{neck}})$), head rotation is constrained by neck motion. This separates intrinsic articulation from extrinsic camera view and gives PPM a stable reference for monocular pose estimation.

\subsubsection{Canonical Injection Strategy.}
Replacing the SMPL-X head with FLAME improves facial fidelity, but the two meshes must connect cleanly at the neck under all poses. Directly stitching meshes in posed space is unstable: small pose errors cause gaps or self-intersections at the neck ring. We instead fuse FLAME and SMPL-T in zero-pose canonical space, where alignment depends only on shape and not on the current pose (\cref{fig:ppm_pipeline}, Right).

The fusion proceeds in four steps:
\begin{enumerate}[leftmargin=*]
    \item \textbf{Head mesh.} Build $V_{\mathrm{head}}(\boldsymbol{\beta}_h, \boldsymbol{\psi}, \boldsymbol{\theta}_{\mathrm{jaw}})$ in canonical space. Only jaw articulation and expression are applied; global head rotation is zero.
    \item \textbf{Torso mesh.} Build $V_{\mathrm{torso}} = \mathbf{J}(\boldsymbol{\beta}_t, \mathbf{0})$ with the SMPL-X torso blendshape function $\mathbf{J}$.
    \item \textbf{Eye-based alignment.} Compute eye-socket centers $C_h$ (head) and $C_p$ (torso) and translate the head:
    \begin{equation}
        V^*_{\mathrm{head}} = V_{\mathrm{head}} - C_h + C_p.
    \end{equation}
    \item \textbf{Injection and skinning.} Replace the SMPL-X head vertices with $V^*_{\mathrm{head}}$, apply neck-ring smoothing $\mathcal{S}(\cdot)$, and drive the unified mesh with LBS:
    \begin{equation}
        V_{\mathrm{out}} = \mathrm{LBS}\!\left(\mathcal{S}(V^*_{\mathrm{head}} \cup V^*_{\mathrm{torso}}),\, \boldsymbol{\theta},\, \mathcal{W}_{\mathrm{unified}}\right).
    \end{equation}
\end{enumerate}

\noindent\textbf{Neck-ring smoothing.}
Let $\mathcal{R}$ denote the fixed set of neck-ring vertices shared by the head and torso templates. For each $i \in \mathcal{R}$, we linearly blend the head and torso positions:
\begin{equation}
    \mathcal{S}(v_i) = (1 - \alpha_i)\, v_i^{\mathrm{head}} + \alpha_i\, v_i^{\mathrm{torso}}, \quad \alpha_i \in [0, 1],
\end{equation}
where $\alpha_i$ increases monotonically from the head side to the torso side of the ring. This produces a watertight transition while preserving FLAME expression and jaw motion on the face region. After fusion, the full portrait mesh is obtained by standard LBS over $\boldsymbol{\theta}$.

\subsection{Progressive Anatomical Alignment}
\label{subsec:paa}

Given PPM, we estimate its parameters with \textbf{Progressive Anatomical Alignment (PAA)}. {PPM exposes the correct portrait variables, but directly regressing all of them remains under-constrained: shoulder keypoints may be noisy, torso geometry is weakly observed, and jaw pose can trade off against expression.} The key idea is to match the anatomical ordering of human motion: global shape and camera first, then skeletal articulation, then non-rigid expression. PAA implements this ordering through both the network architecture and the loss design.

\begin{figure*}[t]
    \centering
    \includegraphics[width=\linewidth]{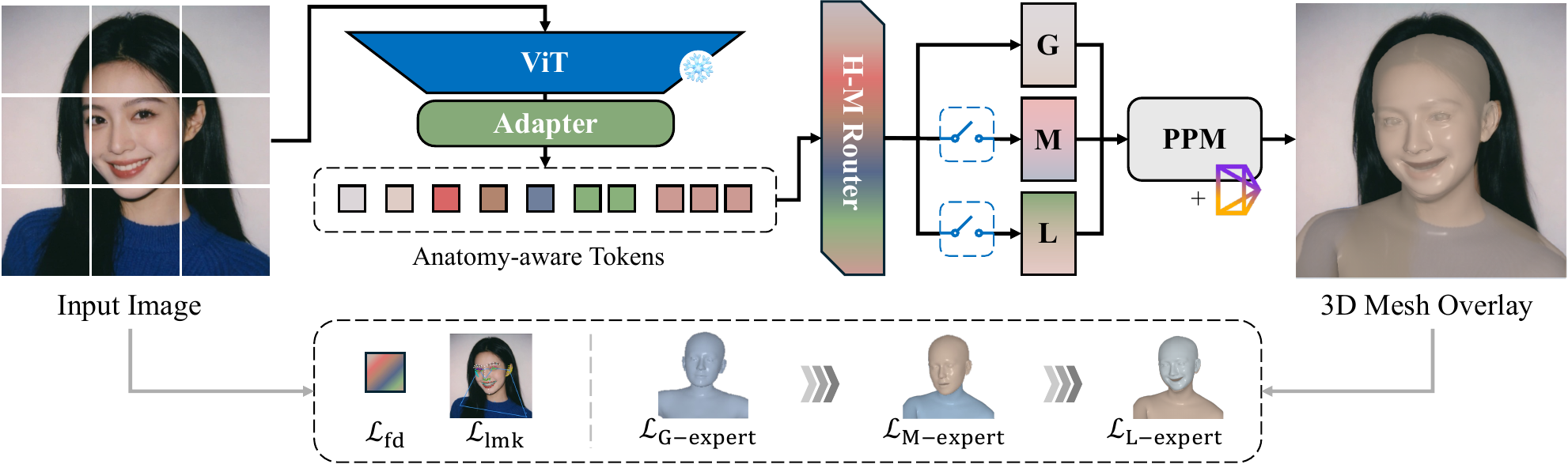}
    \caption{\textbf{Progressive Anatomical Alignment (PAA).}
    {A frozen pretrained encoder extracts image features; a learnable adapter maps them to anatomy-aware tokens.
    Each expert has a learnable query embedding that cross-attends to these tokens (keys/values) to predict the corresponding PPM subset.} The Graduated-Mask Router gates M- and L-expert losses during training.}
    \label{fig:paa_architecture}
\end{figure*}

As shown in \cref{fig:paa_architecture}, the Hierarchical Kinematic Disentanglement-based Expressive (HKD-Exp) Network has three stages:
\begin{enumerate}[leftmargin=*]
    \item A pretrained ViT-based feature extractor extracts portrait features from the input image.
    \item {A learnable adapter converts encoder features into anatomy-aware tokens. Each expert uses its own learnable embedding as the query and these tokens as keys/values in cross-attention, then an MLP regresses the corresponding body-part PPM parameters.}
    \item The predicted parameters are passed through PPM to obtain $\mathbf{V}_{\mathrm{out}}$, which is projected to the image plane for 2D/3D supervision.
\end{enumerate}

\subsubsection{Anatomy-aware Token Extraction.}
Since a global CLS token mixes shape, pose, and expression in one representation, which makes disentanglement harder for downstream regressors, we use a query-based Transformer decoder rather than a single global token.
The input image $\mathbf{I}$ is encoded by a frozen Sapiens-2 encoder~\cite{sapiens2} into spatial features $\mathbf{Z}_{\mathrm{img}} \in \mathbb{R}^{N \times D}$. Sapiens-2 is pretrained on large-scale human images and provides strong portrait features.
{We then apply a learnable feature adapter to obtain anatomy-aware tokens $\mathbf{T}_{\mathrm{anat}} \in \mathbb{R}^{N \times D}$, which serve as the keys and values for expert-wise cross-attention.
For each expert level $k \in \{G, M, L\}$, we introduce a dedicated learnable query embedding $\mathbf{q}_k$. The expert aggregates the corresponding body-part cues by attending to $\mathbf{T}_{\mathrm{anat}}$:}
\begin{equation}
    \mathbf{z}_k = \mathrm{CrossAttn}(\mathbf{q}_k, \mathbf{T}_{\mathrm{anat}}), \quad k \in \{G, M, L\}.
\end{equation}
{In this formulation, $\mathbf{q}_k$ is the query of the $k$-th expert and $\mathbf{T}_{\mathrm{anat}}$ provides keys and values.}
$\mathbf{z}_G$ aggregates information for global shape and camera; $\mathbf{z}_M$ focuses on neck, head, and jaw joints; $\mathbf{z}_L$ focuses on expression and eye details. This splits the regression problem at the feature level before any parameter is predicted.

\subsubsection{Graduated-Mask Hierarchical Experts.}
{The attended feature $\mathbf{z}_k$ is passed to a dedicated MLP expert that predicts the PPM parameters of the corresponding anatomical part:}
1) \textbf{G-Expert ($\mathbf{z}_G$)} predicts camera $\boldsymbol{\pi}$, head shape $\boldsymbol{\beta}_h$, torso shape $\boldsymbol{\beta}_t$. 2) \textbf{M-Expert ($\mathbf{z}_M$)} predicts $\boldsymbol{\theta}_{\mathrm{spine}}$, $\boldsymbol{\theta}_{\mathrm{neck}}$, $\boldsymbol{\theta}_{\mathrm{head}}$, $\boldsymbol{\theta}_{\mathrm{jaw}}$. 3) \textbf{L-Expert ($\mathbf{z}_L$):} expression $\boldsymbol{\psi}$, eye gaze $\boldsymbol{\theta}_{\mathrm{eye}}$.
The predicted parameters are fed into PPM to produce $\mathbf{V}_{\mathrm{out}}$, and the mesh is projected with $\boldsymbol{\pi}$ for loss computation.

\noindent\textbf{Graduated-Mask Router.}
During training, we randomly disable supervision on deeper experts so that each level must remain useful on its own. Let $m_k \in \{0,1\}$ be the mask for level $k$, with $m_G = 1$ always:
\begin{align}
    m_M &= m_G \cdot b_M, \quad b_M \sim \mathrm{Bernoulli}(p), \\
    m_L &= m_M \cdot b_L, \quad b_L \sim \mathrm{Bernoulli}(p).
\end{align}
When $m_L=0$ but $m_M=1$, the network must explain the image with shape and articulation alone; expression cannot absorb jaw errors. This reduces the common shortcut where expression blendshapes mimic mouth opening. At test time, all masks are set to $1$ and all experts are active.

\subsubsection{Training Objectives.}
In-the-wild images lack 3D ground truth. We therefore train PAA with multi-source supervision. Specifically, Sapiens-2~\cite{sapiens2} and 68-point landmarks~\cite{zhu2016face} provide sparse 2D anatomical anchors; Pixel3DMM~\cite{giebenhain2025pixel3dmm} feature distillation transfers representation-level 3D priors; the overflow mask constrains the projected torso extent; and relative geometry preserves the head--torso layout when shoulder detections are missing or noisy. We fit pseudo-labels offline with the Graduated-Refinement Pipeline (\cref{sec:supp_data_pipeline}) and optimize
\begin{equation}
\label{eq:total_loss}
\begin{split}
    \mathcal{L} =\;& \mathcal{L}_{\mathrm{ord}} + \mathcal{L}_{\mathrm{fd}}
    + \underbrace{ \mathcal{L}_{\mathrm{lmk}} + \mathcal{L}_{\mathrm{reg}}^{G}}_{\text{G-Expert}}
    + m_M \underbrace{\left(\mathcal{L}_{\mathrm{nc}} + \mathcal{L}_{\mathrm{om}} + \mathcal{L}_{\mathrm{rg}} + \mathcal{L}_{\mathrm{reg}}^{M}\right)}_{\text{M-Expert}} \\
    &+ m_L \underbrace{\left(\mathcal{L}_{\mathrm{lmk}}^{\mathrm{mouth}} + \mathcal{L}_{\mathrm{lmk}}^{\mathrm{eye}} + \mathcal{L}_{\mathrm{lmk}}^{\mathrm{pupil}} + \mathcal{L}_{\mathrm{reg}}^{L}\right)}_{\text{L-Expert}}.
\end{split}
\end{equation}
Each term is gated by the expert level it supervises. This assigns each supervision source to the anatomical level where it is most useful: global landmarks and shape supervise the G-Expert, anatomical distance and torso geometry supervise the M-Expert, and fine facial landmarks supervise the L-Expert. For simplicity, we omit the weight of each loss term without loss of generality. We describe each loss term below.

\noindent\textbf{Graduated parameter loss ($\mathcal{L}_{\mathrm{ord}}$).}
We apply L2 regression to pseudo-label parameters, with masks matching the Graduated-Mask Router:
\begin{equation}
    \mathcal{L}_{\mathrm{ord}} = \mathcal{L}_{G}(\boldsymbol{\pi}, \boldsymbol{\beta}_h, \boldsymbol{\beta}_t) + m_M \mathcal{L}_{M}(\boldsymbol{\theta}_{\mathrm{spine}}, \boldsymbol{\theta}_{\mathrm{neck}}, \boldsymbol{\theta}_{\mathrm{head}}, \boldsymbol{\theta}_{\mathrm{jaw}}) + m_L \mathcal{L}_{L}(\boldsymbol{\psi}, \boldsymbol{\theta}_{\mathrm{eye}}).
\end{equation}
Each $\mathcal{L}_{G}$, $\mathcal{L}_{M}$, $\mathcal{L}_{L}$ is a standard L2 distance to the corresponding pseudo-label subset.

\noindent\textbf{Feature distillation loss ($\mathcal{L}_{\mathrm{fd}}$).}
We use channel-wise distillation to align student and teacher feature distributions rather than raw feature values. Let $\hat{\mathbf{Z}}, \mathbf{Z} \in \mathbb{R}^{N \times C}$ denote student and teacher token features (per sample), where $N$ is the token dimension and $C$ is the channel dimension. Following~\cite{giebenhain2025pixel3dmm}, we apply temperature-scaled softmax along the token dimension:
\begin{equation}
    \mathbf{P}_s = \log \mathrm{Softmax}\!\left(\frac{F(\hat{\mathbf{Z}})}{\tau}\right), \quad
    \mathbf{P}_t = \mathrm{Softmax}\!\left(\frac{\mathbf{Z}}{\tau}\right),
\end{equation}
and minimize the normalized KL divergence:
\begin{equation}
    \mathcal{L}_{\mathrm{fd}} = \frac{\tau^2}{NC}\,\mathrm{KL}\!\left(\mathbf{P}_s \,\|\, \mathbf{P}_t\right).
\end{equation}
Here $F(\cdot)$ is a 3-layer MLP projection head and $\tau$ is the distillation temperature. This objective transfers teacher attention over tokens and improves anatomy-aware tokens generalization for 3D portrait features.

\noindent\textbf{Global 2D landmark loss ($\mathcal{L}_{\mathrm{lmk}}$).}
Let $\mathbf{u}_j \in \mathbb{R}^2$ denote the $j$-th detected landmark and $\hat{\mathbf{u}}_j$ its projection from PPM. The G-Expert loss combines 68-point facial landmarks~\cite{zhu2016face} (excluding the jawline, indices $0$--$16$) and Sapiens-2 face/shoulder keypoints~\cite{sapiens2}. We use the mean squared error:
\begin{equation}
    \mathcal{L}_{\mathrm{lmk}} = \frac{1}{|\mathcal{J}_G|} \sum_{j \in \mathcal{J}_G} \left\| \hat{\mathbf{u}}_j - \mathbf{u}_j \right\|_2^2,
\end{equation}
where $\mathcal{J}_G$ indexes the supervised landmark set. This term anchors global face placement and head--shoulder layout.

\noindent\textbf{Anatomical distance loss ($\mathcal{L}_{\mathrm{nc}}$).}
Head pose is hard to supervise directly from sparse 2D points. {For jaw disentanglement, we use the nose-tip-to-chin-tip distance as a weak skeletal opening cue: under relatively stable head pose, it reflects jaw-driven chin motion more directly than lip-contour distances, which can be explained by local expression blendshapes.}
\begin{equation}
    \mathcal{L}_{\mathrm{nc}} = \left| d(\hat{\mathbf{u}}_{\mathrm{nose}}, \hat{\mathbf{u}}_{\mathrm{chin}}) - d(\mathbf{u}_{\mathrm{nose}}^{\mathrm{gt}}, \mathbf{u}_{\mathrm{chin}}^{\mathrm{gt}}) \right|,
\end{equation}
where $d(\cdot,\cdot)$ is the Euclidean distance in the image plane.

\noindent\textbf{Overflow mask loss ($\mathcal{L}_{\mathrm{om}}$).}
Torso vertices receive weaker 3D supervision than the face. We therefore penalize projected mesh regions that fall outside the portrait foreground mask $\mathcal{M}_{\mathrm{fg}}$:
\begin{equation}
    \mathcal{L}_{\mathrm{om}} = \frac{\left| \Pi(\mathbf{V}_{\mathrm{torso}}) \setminus \mathcal{M}_{\mathrm{fg}} \right|}{\left| \mathcal{M}_{\mathrm{fg}} \right|},
\end{equation}
where $\Pi(\cdot)$ is mesh rasterization in the image plane and $|\cdot|$ denotes pixel area. This keeps the body mesh inside the visible portrait region. \cref{fig:overfull_mask_viz} shows examples of the overflow mask loss.
% 具体的分析和可视化请参考附录。

\noindent\textbf{Relative geometry loss ($\mathcal{L}_{\mathrm{rg}}$).}
% 我们发现即便加入了带肩部关键点的sapients的pseudo landmark标注，在训练过程中，由于大部分图片的肩部没能完全露出来，导致pseudo landmark的标注缺失或误差明显，导致模型在训练过程中，会使得模型预测的mesh的头部和身体部分的相对关系和比例发生改变，因此我们提出了相对几何关系损失函数，来使得模型预测的mesh的头部和身体部分的比例关系保持不变。同时也能提供头-躯干的相对关系，为head pose和neck pose提供监督。
To add direct 3D supervision on the torso, we compare predicted and pseudo-label meshes in the camera frame after removing global translation. Let $\mathcal{V}_{\mathrm{torso}}$ be the set of mesh vertices outside the injected FLAME head region (torso and shoulders). Denote by $\tilde{\mathbf{v}} = \mathbf{v} - \mathbf{t}$ the vertex position after subtracting the mesh translation. Then
\begin{equation}
    \mathcal{L}_{\mathrm{rg}} = \frac{1}{|\mathcal{V}_{\mathrm{torso}}|} \sum_{\mathbf{v} \in \mathcal{V}_{\mathrm{torso}}} \left\| \tilde{\mathbf{v}}_{\mathrm{pred}} - \tilde{\mathbf{v}}_{\mathrm{gt}} \right\|_2^2.
\end{equation}
Excluding the FLAME head vertices avoids letting face fitting errors dominate the torso loss.

\noindent\textbf{Detail landmark losses.}
The L-Expert uses three landmark groups on fine facial regions:
\begin{equation}
    \mathcal{L}_{\mathrm{lmk}}^{\mathrm{group}} = \frac{1}{|\mathcal{J}_{\mathrm{group}}|} \sum_{j \in \mathcal{J}_{\mathrm{group}}} \left\| \hat{\mathbf{u}}_j - \mathbf{u}_j \right\|_2^2, \quad \mathrm{group} \in \{\mathrm{mouth}, \mathrm{eye}, \mathrm{pupil}\}.
\end{equation}
These terms supervise mouth shape, eye contours, and pupil locations.

\noindent\textbf{Regularization.}
$\mathcal{L}_{\mathrm{reg}}^{G}$, $\mathcal{L}_{\mathrm{reg}}^{M}$, and $\mathcal{L}_{\mathrm{reg}}^{L}$ are L2 penalties on the parameters predicted by each expert, which stabilizes regression when pseudo-labels are noisy.

% \noindent\textbf{Feature-space distillation.}
% Pseudo-labels alone do not cover all in-the-wild appearance variation. We therefore align pooled decoder features with a frozen Pixel3DMM encoder~\cite{giebenhain2025pixel3dmm}:
% \begin{equation}
%     \mathcal{L}_{\mathrm{fd}} = 1 - \cos\!\left(\mathcal{F}_{\mathrm{dec}}(\mathbf{I}),\, \mathcal{F}_{\mathrm{pix}}(\mathbf{I})\right).
% \end{equation}
% $\mathcal{L}_{\mathrm{fd}}$ transfers image-level 3D surface cues in feature space, while $\mathcal{L}_{\mathrm{rg}}$ and $\mathcal{L}_{\mathrm{ord}}$ enforce consistency with pseudo-label geometry. Together they form a dual-space training strategy: representation-space distillation for generalization, and geometry-space alignment for metric portrait structure.

\subsection{Implementation Details}
\label{subsec:impl}

\noindent\textbf{Data preparation.}
We fit PPM parameters offline as pseudo-labels (\cref{sec:supp_data_pipeline}). For each training image, we also extract 68-point landmarks~\cite{zhu2016face}, Sapiens-2 keypoints (face and shoulder), a foreground segmentation mask, and SMPL-X body parameters. These signals supervise different expert levels in \cref{eq:total_loss}.

\noindent\textbf{Two-phase training.}
% We use a frozen Sapiens-2 encoder with PAA (\textcolor{myblue}{learnable anatomy-aware tokens as keys/values, three expert-specific learnable query embeddings, and a 4-layer cross-attention decoder}), as described in \cref{subsec:paa}.
Training has two stages aligned with the graduated design. In stage~1, we pretrain on the multi-view Nersemble dataset~\cite{kirschstein2023nersemble} to learn stable shape and camera estimation; both $m_M$ and $m_L$ are always $1$. In stage~2, we finetune on LS3DW~\cite{bulat2017far}, CelebA~\cite{liu2015faceattributes}, LaPa~\cite{liu2020new}, and LFW~\cite{huang2007lfw} with graduated masking enabled ($p=0.5$) so that the router regularizes jaw--expression disentanglement on in-the-wild data. All experiments use 4 NVIDIA A100 GPUs, Adam optimizer, batch size 64, and 4M iterations ($\sim$5 days). Additional hyperparameters are in the appendix (\hyperref[sec:supp_impl]{Implementation Details}).

% !TEX root = ../main.tex
\section{Experiments}
\label{sec:exp}

{We organize the experiments around five research questions:}
\begin{itemize}
    \item \textbf{RQ1 (Geometric reconstruction):}
    Does \methodname{} improve in-the-wild landmark accuracy and 3D head mesh fitting compared to recent monocular baselines?
    \item \textbf{RQ2 (Jaw--expression disentanglement):}
    On talking videos, does graduated routing yield consistent jaw motion and stable temporal dynamics?
    \item \textbf{RQ3 (Head--shoulder kinematics):}
    Does PPM with PAA improve head--torso coherence?
    \item \textbf{RQ4 (Component contribution):}
    {How much does each part of the design contribute, including loss functions, data augmentation, router activation $\alpha_{\mathrm{act}}$, and the architecture?}
    \item \textbf{RQ5 (Downstream application):}
    {Does \methodname{} benefit downstream tasks such as talking-head generation and 3D portrait generation?}
\end{itemize}

\subsection{Experimental Setup}
\label{subsec:setup}

\textbf{Datasets.}
{We follow the two-phase training in \cref{subsec:impl}.}
We pretrain on the multi-view \dsNer{}~\cite{kirschstein2023nersemble} training set and finetune on \dsLsThreeDw~\cite{bulat2017far}, \dsCelebA~\cite{liu2015faceattributes}, \dsLaPa~\cite{liu2020new}, and \dsLfw~\cite{huang2007lfw}, using PPM pseudo-labels from our Graduated-Refinement Pipeline (details in \cref{sec:supp_data_pipeline}).
{We evaluate under three protocols:}
(1)~\emph{image landmarks} on the test set of four image benchmarks above;
(2)~\emph{talking-head video} on \dsHdtf{}~\cite{zhang2021flow}, which contains 20 randomly selected videos;
and (3)~\emph{3D reconstruction} on the \dsNer{}~\cite{kirschstein2023nersemble} test set, NoW~\cite{sanyal2019learning}, and Stirling~\cite{feng2018evaluation}.
{Image benchmarks report 68-point Normalized Mean Error (NME), excluding the jawline (indices 0--16)~\cite{wu2026pear}.}
{Video benchmarks report the jaw--skeletal ratio score (JSR) and temporal stability.}
{On \dsNer{}, we report mesh vertex error (MVE/LVE)~\cite{feng2021learning}.}
{NoW and Stirling report mean/median/std between the registered mesh and the ground-truth mesh with their official code.}
Metric definitions and protocol notes are in the supplement (\cref{sec:exp_supp}).

\textbf{Baselines.}
{We compare against DECA~\cite{feng2021learning}, EMOCA~\cite{danve2022emoca}, Deep3DFace~\cite{deng2019accurate}, TEASER~\cite{Liu2025TEASER}, SPECTRE~\cite{filntisis2022visual}, 3DDFA-V2~\cite{guo2020towards}, 3DDFA-V3~\cite{wang20243d}, SMPLest-X~\cite{yin2025smplest}, PEAR~\cite{wu2026pear}, and SAM-3DB~\cite{yang2026sam3dbody}.}
{Each method uses its own official crop for fair comparison.}
{For visualization, we paste the reconstructed mesh back onto the original image for all methods.}

\subsection{Qualitative Comparison}
\label{subsec:qual}

\begin{figure}[htbp]
    \centering
    \includegraphics[width=\linewidth]{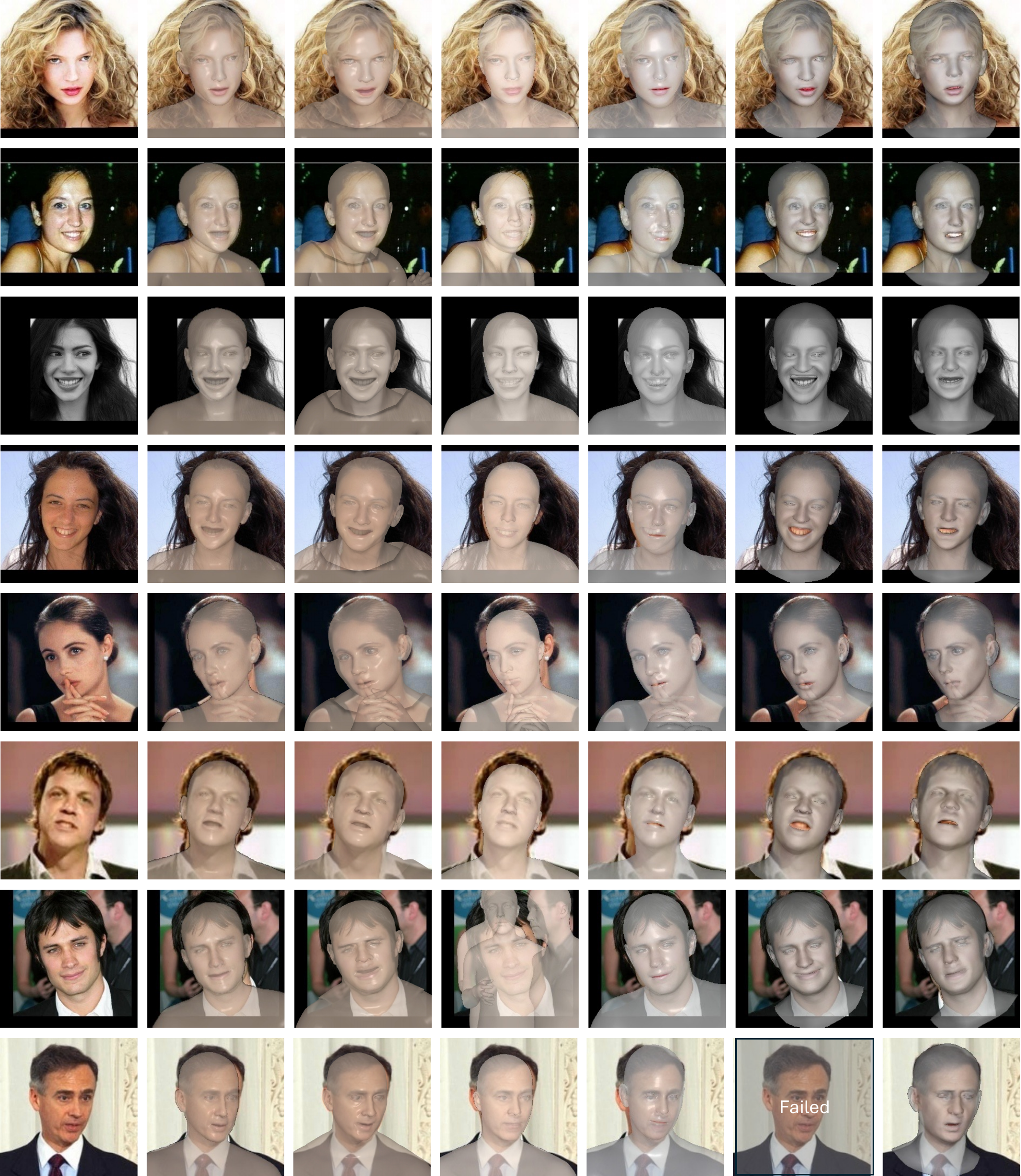}\\[-0.4em]
    {\footnotesize
    \noindent
    \begin{tabular}{@{}*{7}{>{\centering\arraybackslash}p{\dimexpr\linewidth/7\relax}@{}}}
      Input
      & Ours
      & PEAR~\cite{wu2026pear}
      & SAM-3DB~\cite{yang2026sam3dbody}
      & SMPLest-X~\cite{yin2025smplest}
      & SPECTRE~\cite{filntisis2022visual}
      & TEASER~\cite{Liu2025TEASER} \\
    \end{tabular}}
    \caption{\textbf{Qualitative reconstruction on in-the-wild images.}}
    \label{fig:qual_recon}
\end{figure}

\cref{fig:qual_recon} compares mesh overlays on in-the-wild portraits with visible shoulders.
{FLAME/BFM-based baselines often detach the head from the torso.}
{\methodname{} recovers a continuous head--neck--shoulder mesh that follows the input.}
{We also observe more stable ear placement than several baselines.}
{PEAR and SAM-3DB can predict plausible shoulders and head pose.}
{However, PEAR often shows inaccurate shape and a clear gap between head and torso.}
{SAM-3DB does not estimate facial expression and often predicts a head that is too large or too small, which causes misalignment.}
This supports \textbf{RQ1} and \textbf{RQ3}.

\begin{figure}[htbp]
    \centering
    \includegraphics[width=\linewidth]{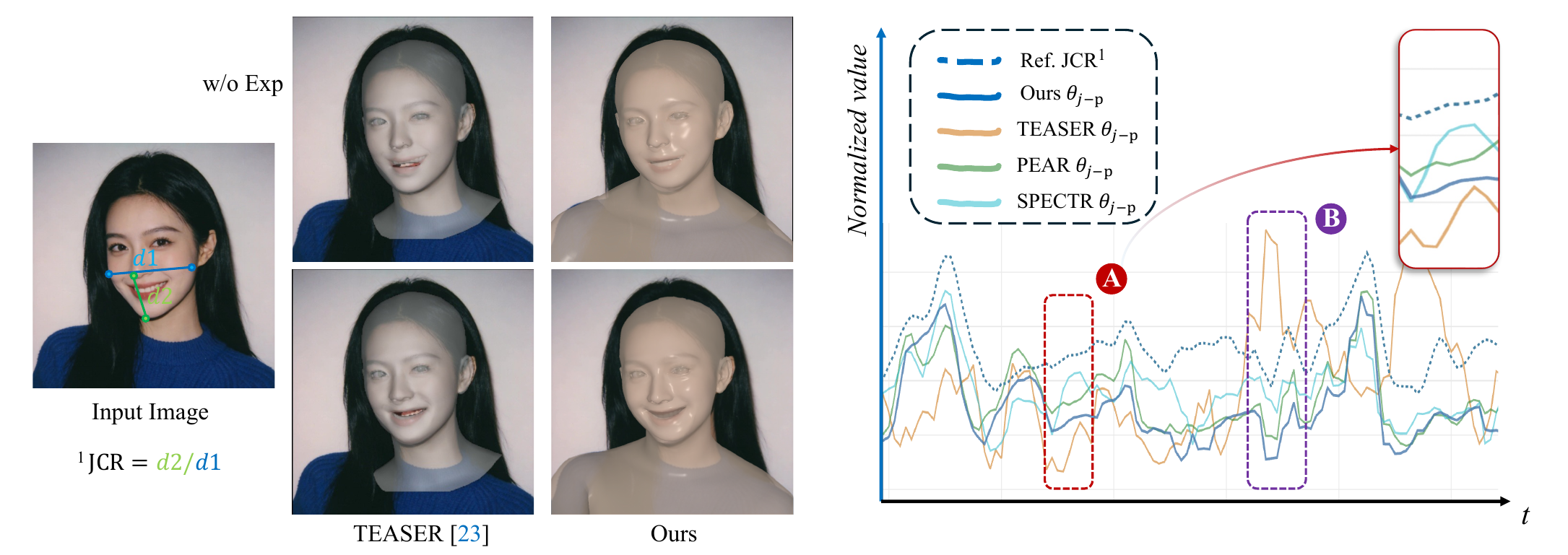}%
    \caption{\textbf{Jaw--skeletal consistency during speech.}
    \textbf{Left:} reconstructions with and without expression on the same talking frame.
    \textbf{Right:} time series of the normalized nose--chin reference JCR (blue) versus predicted jaw pitch for \methodname{} (beige) and TEASER (black).
    The highlighted interval shows TEASER attributing mouth motion to expression while jaw pitch fails to track $JCR$.}
    \label{fig:jaw_distance_analysis}
\end{figure}

\begin{figure}[htbp]
    \centering
    \begin{tikzpicture}[inner sep=0pt, outer sep=0pt]
      \node[anchor=south west] (img) {%
        \includegraphics[width=\dimexpr\linewidth-1.1em\relax]{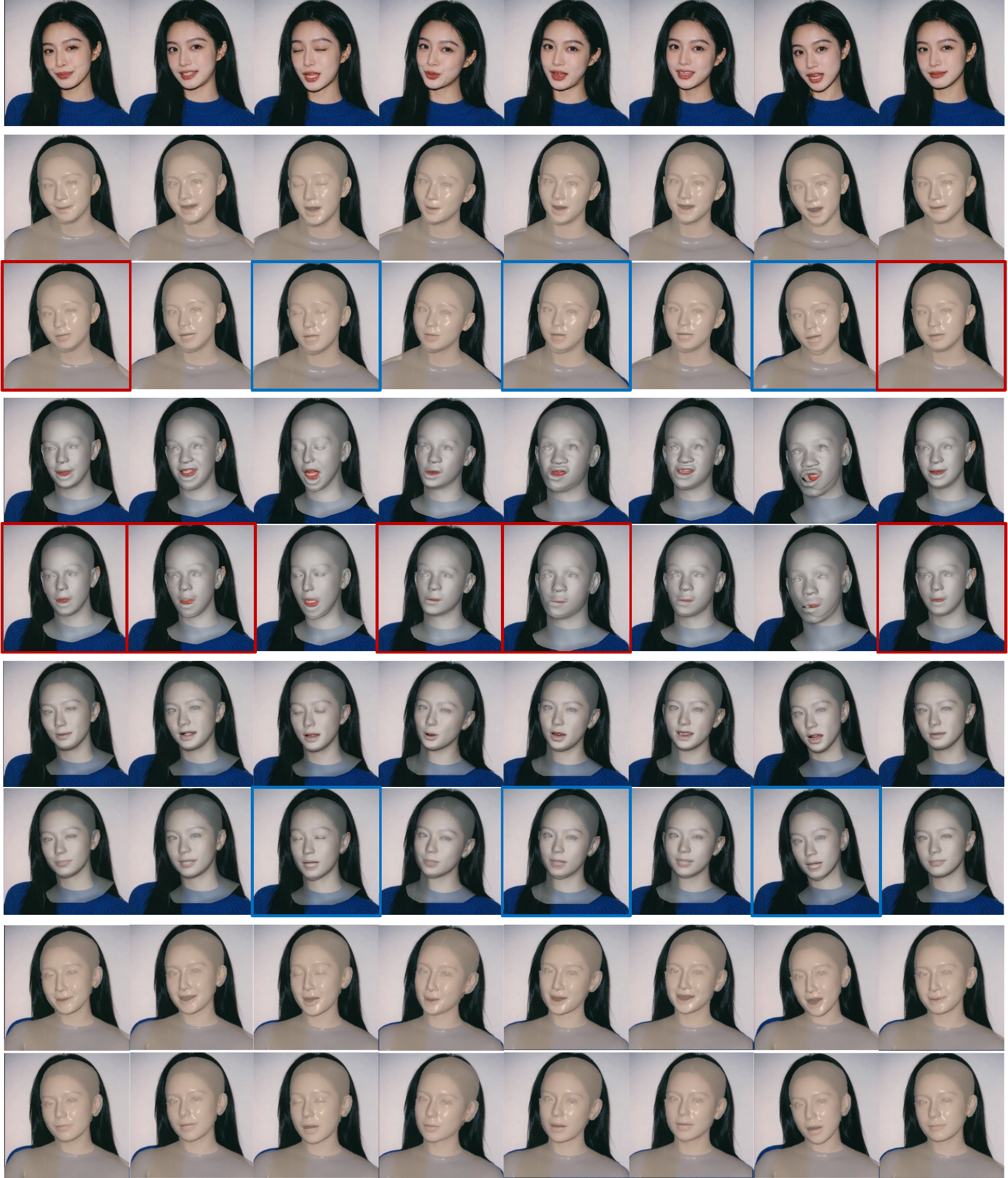}};
      % 9 rows from top: Input(1) + PEAR/SPECTRE/TEASER/Ours (2 each).
      % Positions are centers along the left edge (bottom $\rightarrow$ top).
      \path (img.south west) -- (img.north west)
        node[pos=1/9,  anchor=south, rotate=90, font=\footnotesize, inner sep=1pt] {Ours}
        node[pos=3/9,  anchor=south, rotate=90, font=\footnotesize, inner sep=1pt] {TEASER~\cite{Liu2025TEASER}}
        node[pos=5/9,  anchor=south, rotate=90, font=\footnotesize, inner sep=1pt] {SPECTRE~\cite{filntisis2022visual}}
        node[pos=7/9,  anchor=south, rotate=90, font=\footnotesize, inner sep=1pt] {PEAR~\cite{wu2026pear}}
        node[pos=17/18, anchor=south, rotate=90, font=\footnotesize, inner sep=1pt] {Input};
    \end{tikzpicture}
    \caption{\textbf{Jaw--expression disentanglement comparison on a talking clip.}
    The \textcolor{deepred}{\textbf{deep-red box}} highlights a case that is anatomically inconsistent:
    the jaw should close the mouth while expression opens it, yet both jaw and expression open the mouth.
    The \textcolor{deepblue}{\textbf{deep-blue box}} highlights a case where the jaw should drive mouth opening,
    but the predicted jaw under-opens and expression dominates.
    }
    \label{fig:disentangle_vis}
\end{figure}

{To answer \textbf{RQ2}, we compare predicted jaw pitch ($\theta_{j-p}$) with skeletal mouth opening during speech.}
{We use the \emph{jaw-chain ratio} (JCR), a normalized nose--chin distance that serves as a skeletal mouth-opening reference.}
{We then report the \emph{jaw--skeletal ratio score} (JSR), the Pearson correlation between predicted jaw pitch and JCR.}
{The left part of \cref{fig:jaw_distance_analysis} shows a case where TEASER~\cite{Liu2025TEASER} keeps jaw pitch entangled with expression.}
{In that frame, the smile is driven by expression and the teeth remain closed.}
{After removing expression, TEASER still predicts a large jaw pitch, so the mouth stays too open relative to the skeletal cue.}
{\methodname{} tracks JCR more closely, which suggests that jaw articulation, not blendshapes, drives the skeletal motion.}
{The right plot of \cref{fig:jaw_distance_analysis} shows that other methods can still produce a plausible face mesh, while their jaw pitch stays weakly correlated with the JCR curve.}
The regions marked by
\circledmark{deepred}{A}
and
\circledmark{deeppurple}{B}
highlight this mismatch.
{\cref{fig:disentangle_vis} shows a high mouth-opening frame from the same clip.}
{Compared with TEASER~\cite{Liu2025TEASER}, PEAR~\cite{wu2026pear}, and SPECTRE~\cite{filntisis2022visual}, \methodname{} assigns large mouth opening more consistently to jaw articulation rather than to expression shortcuts.}

\begin{figure}[htbp]
  \centering
  \includegraphics[width=\linewidth]{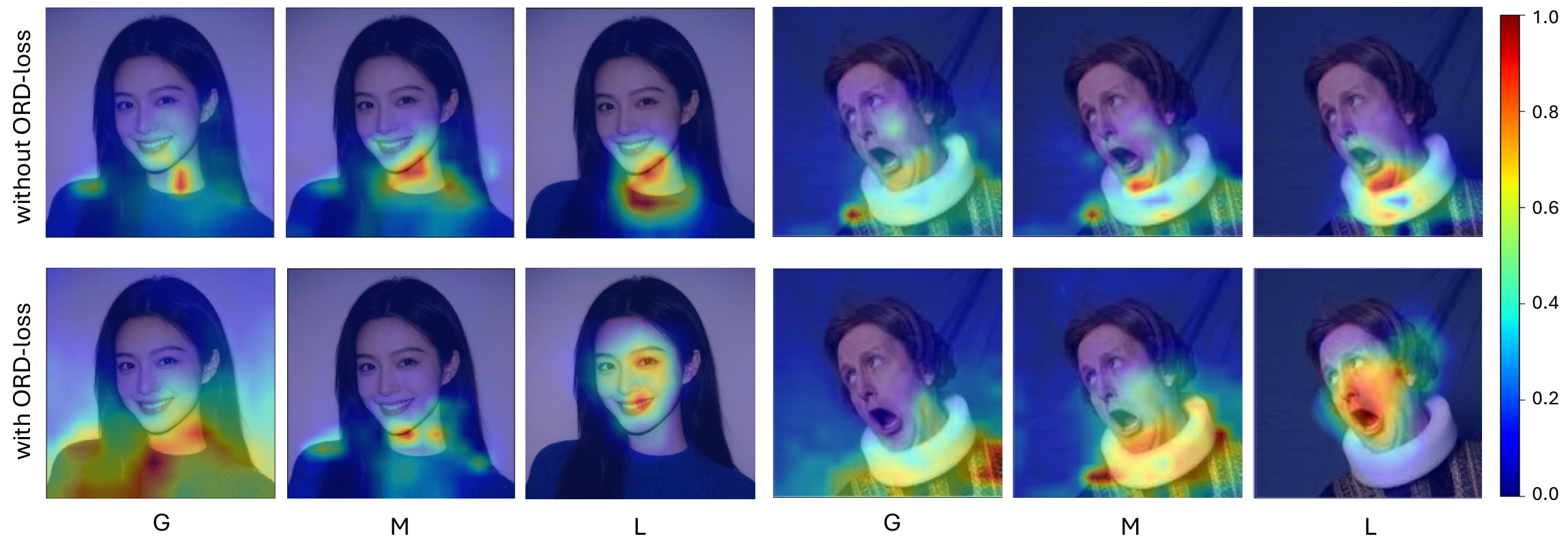}%
  \caption{\textbf{Expert-net attention visualization.}
  With $\mathcal{L}_{\mathrm{ord}}$, our expert nets focus on the accurate regions of the input image.}
  \label{fig:attention_viz}
\end{figure}

\subsection{Quantitative Comparison}
\label{subsec:quant}

{We answer \textbf{RQ1--RQ3} with image, video, and 3D benchmarks.}

\textbf{Image landmark accuracy.}
\cref{tab:cmp_landmark} reports 68-point NME on four in-the-wild datasets.
{\methodname{} achieves the best average NME ($7.81$) among the reported methods.}
{TEASER~\cite{Liu2025TEASER} is second on average ($10.72$) but does not model the torso.}
{3DDFA-V3~\cite{wang20243d} has higher NME under the reported protocol.}

% TODO: verify baseline numbers against current eval script.
\begin{table}[t]
  \centering
  \caption{{Projection alignment comparison in image space. We report 68-keypoints-based normalized mean error (NME $\times 100$) excluding the jawline keypoints.
    Each method uses its own official crop protocol for fair comparison.}}
  \label{tab:cmp_landmark}
  \small
  \setlength{\tabcolsep}{3.5pt}
  \begin{tabular}{lcccccc}
    \toprule
    Method
      & 3DMM Basis
      & \dsLsThreeDw\ \mdown
      & \dsCelebA\ \mdown
      & \dsLaPa\ \mdown
      & \dsLfw\ \mdown
      & Avg.\ \mdown \\
    \midrule
    % AUTO_START tab_cmp_landmark
    DECA~\cite{feng2021learning} & FLAME2020 & 23.88 & 21.60 & 25.99 & 20.01 & 22.87 \\
    EMOCA~\cite{danve2022emoca} & FLAME2020 & 31.03 & 22.49 & 29.99 & 19.98 & 28.87 \\
    SPECTRE~\cite{filntisis2022visual} & FLAME2020 & 58.51 & 46.32 & 51.21 & 45.70 & 50.44 \\
    3DDFA-V3~\cite{wang20243d} & BFM & 18.42 & 13.66 & 15.95 & 13.65 & 15.42 \\
    TEASER~\cite{Liu2025TEASER} & FLAME2020 & \second{15.72} & \second{7.01} & \second{9.88} & 10.26 & \second{10.72} \\
    SMPLest-X~\cite{yin2025smplest} & SMPL-X & 91.79 & 79.99 & 71.53 & 80.29 & 78.40 \\
    PEAR~\cite{wu2026pear} & EHM-s & 19.24 & 8.74 & 11.87 & \second{10.02} & 12.46 \\
    \methodname{} (Ours) & PPM & \best{10.56} & \best{5.63} & \best{7.07} & \best{8.22} & \best{7.81} \\
    % AUTO_END tab_cmp_landmark
    \bottomrule
  \end{tabular}
\end{table}

\textbf{Talking-head motion.}
\cref{tab:cmp_hdtf} evaluates talking sequences on \dsHdtf.
{\methodname{} obtains the best JSR ($0.799$) among compared methods.}
{It also reports the best NME-P ($0.058$) and NME-S ($0.155$).}
{Temporal stability is second-best ($1.680$), after SMPLest-X~\cite{yin2025smplest}, which has limited expression dimension (10 vs. 50) and less-expressive mouth movements.}

% Prefer JSR for disentanglement; Var-Exp/Var-Jaw highlight unbalanced dynamics.
\begin{table}[t]
  \centering
  \caption{{Facial motion evaluation on the \dsHdtf{} test set ($20$ randomly selected videos).
    \textbf{JSR}: Pearson $r$ between predicted jaw pitch and the jaw-chain ratio (JCR), a normalized nose--chin distance reference (see \cref{fig:jaw_distance_analysis}); higher means jaw articulation tracks skeletal motion rather than expression shortcuts.
    \textbf{Var-Exp} / \textbf{Var-Jaw}: temporal variance of expression and jaw parameters.
    \textbf{Temporal}: mean second-order difference of jaw and expression ($\downarrow$ smoother).
    Each method uses its own official crop protocol.
    In the Var-Exp / Var-Jaw columns, \textcolor{deepred}{\textbf{deep-red}} marks abnormally high values among baselines, and \textcolor{deepblue}{\textbf{deep-blue}} marks abnormally low values (same color coding as \cref{fig:disentangle_vis}).}}
  \label{tab:cmp_hdtf}
  \small
  \setlength{\tabcolsep}{6pt}
  \begin{tabular}{lcccccc}
    \toprule
    Method & JSR \mup & (Var-Exp & Var-Jaw) & Temporal \mdown & NME-P \mdown & NME-S \mdown \\
    \midrule
    % AUTO_START tab_cmp_hdtf
    SPECTRE~\cite{filntisis2022visual} & 0.564 & \textcolor{deepblue}{0.104} & \textcolor{deepred}{0.601} & 1.847 & 0.179 & -- \\
    TEASER~\cite{Liu2025TEASER} & 0.371 & \textcolor{deepred}{0.490} & {0.197} & 2.344 & \second{0.066} & -- \\
    SMPLest-X~\cite{yin2025smplest} & 0.591 & \textcolor{deepblue}{0.003} & {0.207} & \best{1.040} & 0.117 & \second{0.260} \\
    PEAR~\cite{wu2026pear} & 0.537 & \textcolor{deepred}{0.414} & \textcolor{deepblue}{0.123} & 1.851 & 0.072 & 0.849 \\
    \methodname{} (Ours) & \best{0.799} & 0.297 & 0.348 & \second{1.680} & \best{0.058} & \best{0.155} \\
    % AUTO_END tab_cmp_hdtf
    \bottomrule
  \end{tabular}
\end{table}

\textbf{3D mesh and pose.}
{\cref{tab:cmp_3d_mesh} reports 3D errors on \dsNer{}, NoW, and Stirling.}
{On \dsNer{}, \methodname{} achieves the best MVE ($8.20$) and LVE ($6.19$) among compared methods.}
{On NoW, \methodname{} has the best mean ($1.09$) and tied-best std ($0.99$), with median second to TEASER~\cite{Liu2025TEASER}.}
{On Stirling, \methodname{} has the best median ($0.99$) and std ($1.03$), with mean second to TEASER.}

% \input{tables/tab_cmp_nersemble_mesh}
% TODO: verify baseline numbers against current eval script.
\begin{table}[t]
  \centering
  \caption{3D head mesh accuracy comparison. We report mean vertex error (MVE $\times 10^3$) and lip vertex error (LVE $\times 10^4$) on Nersemble-V2, and mean/median/std on NoW and Stirling.}
  \label{tab:cmp_3d_mesh}
  \small
  \setlength{\tabcolsep}{4pt}
  \begin{tabular}{l|cc|ccc|ccc}
    \toprule
    & \multicolumn{2}{c|}{Nersemble-V2} & \multicolumn{3}{c|}{NoW benchmark} & \multicolumn{3}{c}{Stirling benchmark} \\
    Method & MVE  \mdown & LVE \mdown & Median \mdown & Mean \mdown & Std  \mdown & Median  \mdown & Mean  \mdown & Std  \mdown  \\
    \midrule
    Deep3DFace$^*$~\cite{deng2019accurate} & -- & -- & 1.11 &  1.41 & 1.21 & \best{0.99} & 1.27 & 1.15 \\
    DECA~\cite{feng2021learning} & 60.67 & 74.72 & 1.23 & 1.57 & 1.39 & 1.03 & 1.32 & 1.18 \\
    EMOCA~\cite{danve2022emoca} & 63.29 & 36.37 & 1.24 & 1.56 & 1.44 & 1.02  & 1.32 & 1.28 \\
    3DDFA-V2~\cite{guo2020towards} & -- & -- & 1.09 & 1.38 & 1.18 & 1.20 & 1.55 & 1.45 \\
    3DDFA-V3~\cite{wang20243d} & -- & -- & 1.05 & 1.33 & 1.20 & 1.19  & 1.56 & 1.43 \\
    SPECTRE~\cite{filntisis2022visual} & 67.92 & 59.32 & 1.66 & 1.37 & 1.27 & 1.04  & 1.19 & 1.25 \\
    SMIRK~\cite{retsinas2024smirk} & 37.88 & 40.19 & 0.99 & 1.22 & 1.02 & 1.01 & 1.08 & 1.05 \\
    TEASER~\cite{Liu2025TEASER} & 35.04 & 32.02 & \best{0.92} & \second{1.10} & \best{0.99} & \second{1.00} & \best{1.07} & \second{1.04} \\
    PEAR~\cite{wu2026pear} & \second{13.62} & \second{8.08} & 0.94 & 1.17 & 1.08 & 1.03  & 1.20 & 1.11 \\
    \methodname{} (Ours) & \best{8.20} & \best{6.19} & \second{0.93} & \best{1.09} & \best{0.99} & \best{0.99}  & \second{1.08} & \best{1.03} \\
    \bottomrule
  \end{tabular}
\end{table}

\subsection{Ablation Studies and Analysis}
\label{subsec:ablation}

{We ablate structural components that match our claims (\cref{tab:ablation_strong}).}
{A flat regressor removes anatomy-aware queries and predicts all PPM parameters from one pooled feature.}
{``w/o Graduated-Mask Router'' keeps the same experts but disables stochastic graduated masking.}
{``w/o graduated pseudo-labels'' trains on jointly fitted pseudo-labels.}
{``w/o PPM torso anchor'' replaces the torso-rooted PPM with a floating-head FLAME output.}
{Relative to the full model, removing PAA drops JSR from $0.799$ to $0.368$ and raises average NME from $7.81$ to $10.44$.}
{Removing the Graduated-Mask Router mainly hurts JSR ($0.411$).}
{Removing graduated pseudo-labels raises average NME to $12.55$ and lowers JSR to $0.701$.}
{Removing the PPM torso anchor raises MPJPE from $69.4$ to $89.1$, while landmark and mesh errors stay close to the full model.}

% TODO: verify final-recipe checkpoints.
\begin{table*}[htbp]
  \centering
  \caption{Structural ablation of the proposed anatomical design.
    Each row removes one high-level component while keeping the remaining training recipe unchanged.
    JSR measures jaw--skeletal correlation on \dsHdtf{}; NME ($\times 100$) measures image-space face alignment; MVE ($\times 10^3$)/LVE ($\times 10^4$) measure 3D head/mouth mesh accuracy; MPJPE measures head--shoulder joint accuracy.}
  \label{tab:ablation_strong}
  \small
  \setlength{\tabcolsep}{4pt}
  \begin{tabular}{lccccc}
    \toprule
    Variant & \dsHdtf{} JSR \mup & Avg. NME \mdown & Ner-MVE \mdown & Ner-LVE \mdown & MPJPE \mdown \\
    \midrule
    Flat regressor w/o PAA & 0.368 & 10.44 & 24.89 & 26.01 & 79.0 \\
    w/o Graduated-Mask Router & 0.411 & \second{8.07} & 8.31 & 6.20 & \second{71.3} \\
    w/o graduated pseudo-labels & 0.701 & 12.55 & 11.92 & 9.53 & 76.6 \\
    w/o PPM torso anchor & \second{0.782} & 8.11 & \best{8.23} & \best{6.15} & 89.1 \\
    \methodname{} full & \best{0.799} & \best{7.81} & \second{8.20} & \second{6.19} & \best{69.4} \\
    \bottomrule
  \end{tabular}
\end{table*}

{Training objectives and Graduated-Mask Router settings for \textbf{RQ4} are reported in the supplement (\cref{tab:ablation_components,sec:exp_supp}).}
{\cref{fig:attention_viz} further visualizes expert attention under the graduated parameter loss $\mathcal{L}_{\mathrm{ord}}$.}
{With this loss, different experts attend to the image regions that match their anatomical roles.}

\subsection{Applications}
\label{subsec:app}

\begin{figure}[htbp]
  \centering
  \includegraphics[width=\linewidth]{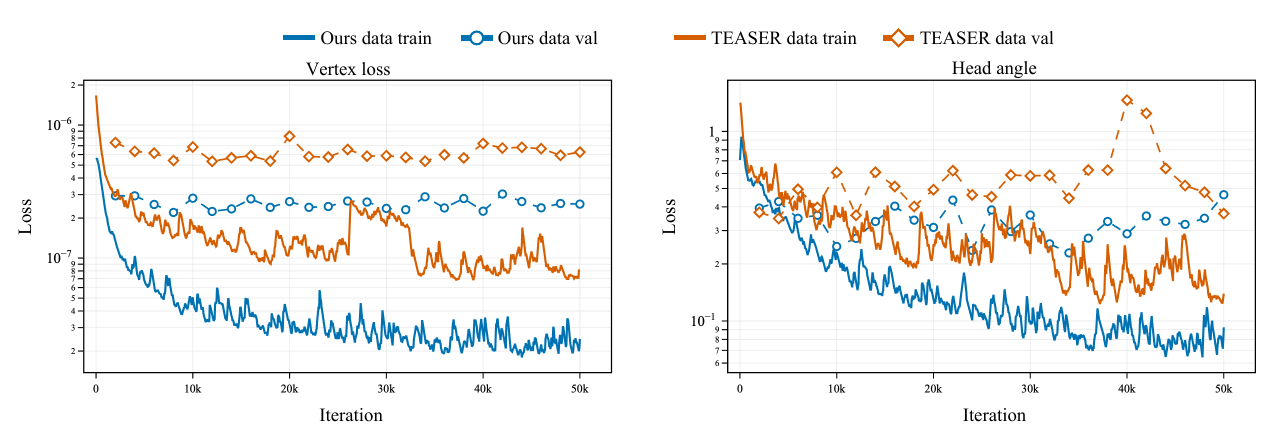}%
  \caption{\textbf{Talking-head training on \dsHdtf{}.}
  We train DiffPoseTalk~\cite{sun2024diffposetalk} on TEASER-generated parameters and on \methodname{}-generated parameters.
  The loss curves show more stable training and faster convergence with the \methodname{}-labeled set.}
  \label{fig:talking_head}
\end{figure}

\paragraph{Audio-driven Talking-Head Generation}

{We test whether reconstructed parameters help downstream talking-head training (\textbf{RQ5}).}
{Following DiffPoseTalk~\cite{sun2024diffposetalk}, we build two training sets from the same \dsHdtf{} clips ({$10$ randomly selected video clips}): one from TEASER~\cite{Liu2025TEASER} parameters and one from \methodname{} parameters.}
{\cref{fig:talking_head} shows the training loss curves.}
The \methodname{}-labeled set yields more stable training and faster convergence.
% \exptodo{add quantitative generation metrics (e.g., lip sync / FID / user study) if available; current claim follows the training curves only.}
{This observation is consistent with the stronger JSR and NME-P/NME-S in \cref{tab:cmp_hdtf}, which indicate cleaner jaw motion and better alignment for animation.}

\begin{figure}[htbp]
  \centering
  \includegraphics[width=\linewidth]{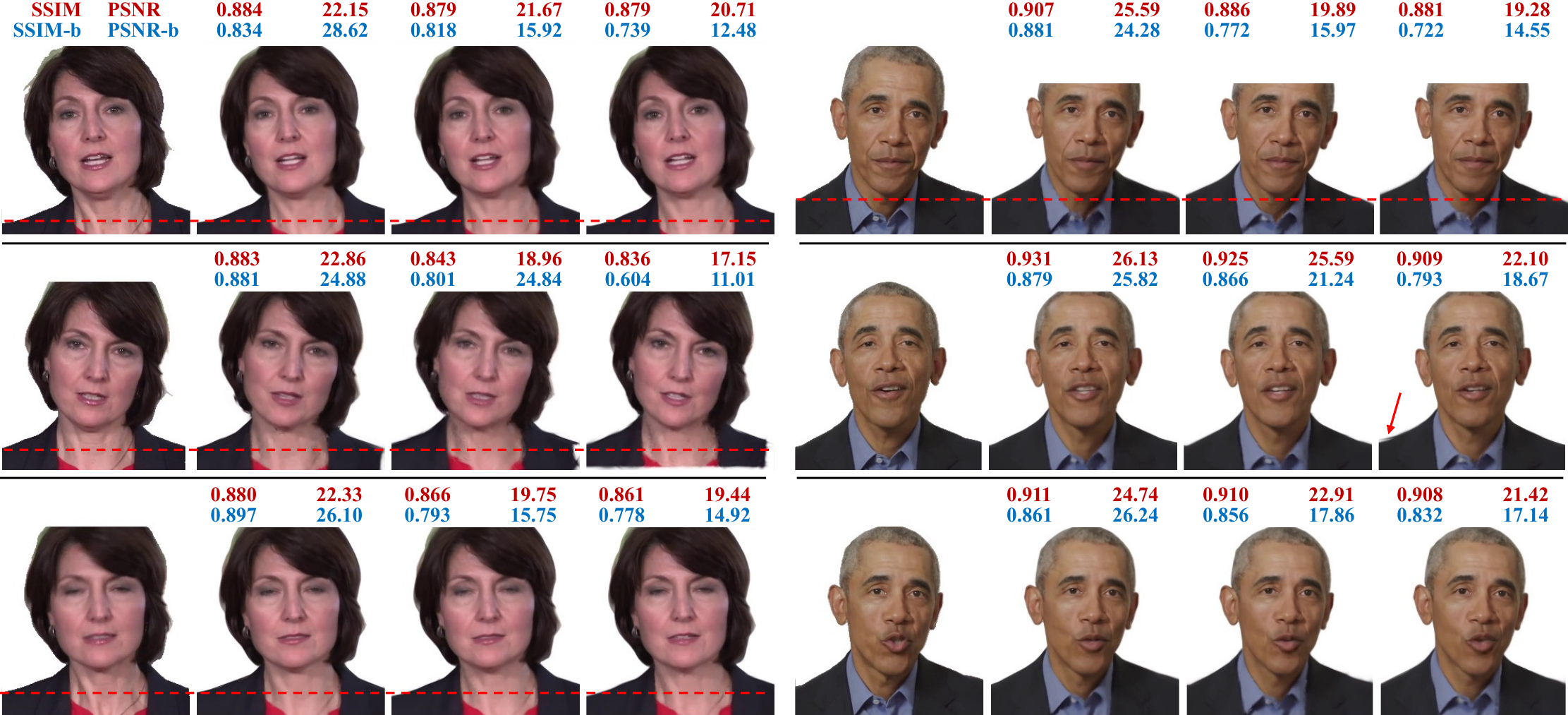}\\[-0.4em]
  {\footnotesize
  \noindent
  \begin{tabular}{@{}
    *{4}{>{\centering\arraybackslash}p{\dimexpr(\linewidth-0.03\linewidth)/8\relax}@{}}
    @{\hspace{0.03\linewidth}}
    *{4}{>{\centering\arraybackslash}p{\dimexpr(\linewidth-0.03\linewidth)/8\relax}@{}}
  @{}}
    Reference
    & Ours
    & Ours (head)
    & TEASER~\cite{Liu2025TEASER}
    & Reference
    & Ours
    & Ours (head)
    & TEASER~\cite{Liu2025TEASER} \\
  \end{tabular}}
  \caption{\textbf{Avatar generation and animation on \dsHdtf{}.}
  Based on RGBAvatar~\cite{li2025rgbavatar}, we compare avatars trained on TEASER parameters, \methodname{} head-only parameters, and full \methodname{} parameters.
  Relative to TEASER, full \methodname{} reduces the ``floating head'' artifact and keeps better continuity in the neck--shoulder region. Here SSIM-b and PSNR-b indicate the SSIM and PSNR of the body part only.}
  \label{fig:avatar_rec}
\end{figure}

\paragraph{Animatable 3D Gaussian Splatting Avatar Generation}

{We further test animatable avatar reconstruction with RGBAvatar~\cite{li2025rgbavatar} (\textbf{RQ5}).}
{We train RGBAvatar on three settings from the same \dsHdtf{} sequences: (1)~TEASER-generated parameters, (2)~\methodname{} head-only parameters (same head mesh topology as TEASER), and (3)~full \methodname{} parameters.}
{\cref{fig:avatar_rec} compares the reconstructed avatars against the reference.}
{The TEASER-based avatar shows clearer artifacts around the shoulders due to the floating-head issue.}
{The full \methodname{}-based avatar keeps a more continuous head--neck--shoulder surface.}
Quantitative SSIM / PSNR results are reported in the supplement (\cref{tab:cmp_rgbavatar}). In short, full \methodname{} parameters outperform TEASER and head-only \methodname{} parameters in terms of both visual quality and quantitative metrics.

% !TEX root = ../main.tex
\section{Conclusion}
\label{sec:conclusion}

We presented \textbf{GRAPE} for monocular portrait mesh estimation. PPM models the torso-to-head kinematic chain and merges FLAME with SMPL-X through canonical injection, which reduces the ambiguity between camera pose and head articulation. PAA regresses PPM parameters in coarse-to-fine order using a Graduated-Mask Router, which helps separate jaw motion from expression blendshapes. Experiments show improved pose alignment and jaw--expression disentanglement, and the output parameters are suitable for talking-head animation.

\noindent\textbf{Limitations and Future Work.}
PPM assumes a tight skin surface, so neck and shoulder reconstruction can fail under loose clothing or heavy neck occlusion; the face region remains more stable. Pseudo-label quality also depends on 2D landmarks, which can be noisy under extreme lighting or profile views. Future work may combine implicit representations (NeRF or Gaussian splatting) with explicit kinematic control to better handle hair and clothing.

% \section*{Acknowledgements}
% Please insert your acknowledgments here.

% Prebuilt bibliography for arXiv (upload main.bbl; do not rely on remote bibtex)
\bibliographystyle{ACM-Reference-Format}

\appendix
% !TEX root = ../main.tex
\newpage
\begin{center}
    {\Large\bfseries Appendix}
\end{center}
\phantomsection\label{sec:supp}
% \label{sec:supp}

This section contains additional details on the implementation of the proposed method. The appendix contains:
\begin{itemize}
    \item \textbf{Implementation details.} This section provides additional details on the implementation of the proposed method, including the choice of hyperparameters and the implementation of the backbones and expert net.
    \item \textbf{Data Generation Pipeline.} This section provides additional details on the data generation pipeline, including the choice of each component techniques and the optimization design.
    \item \textbf{Additional Experimental Results.} This section provides additional experimental results, including more results of the proposed method on image and video inputs.
    \item \textbf{Limitations, Discussions and Future Work.} This section provides additional discussions on the limitations of the proposed method, and suggests future work.
\end{itemize}

\section{Implementation Details}
\label{sec:supp_impl}

This section complements \cref{subsec:impl}. Portrait crops are resized to $256 \times 256$. The cross-attention decoder has 4 layers with hidden size $D{=}768$. Loss weights for $\mathcal{L}_{\mathrm{ord}}$, $\mathcal{L}_{\mathrm{lmk}}$, $\mathcal{L}_{\mathrm{nc}}$, $\mathcal{L}_{\mathrm{om}}$, $\mathcal{L}_{\mathrm{rg}}$, and $\mathcal{L}_{\mathrm{fd}}$ are chosen on a held-out validation set. During NersembleV2~\cite{kirschstein2023nersemble} pretraining, we set $m_M = m_L = 1$ (equivalently $p{=}1.0$); during in-the-wild finetuning, the Graduated-Mask Router uses $p{=}0.5$. We train with Adam, batch size 64, for 4M iterations on 4 NVIDIA A100 GPUs.

\subsection{PPM Construction Details}
\label{subsec:supp_ppm_construction}

PPM is implemented as a single skinned portrait template whose face region follows FLAME2023 and whose neck--shoulder region follows the upper-body part of SMPL-X. The construction is performed once in canonical space and reused during both pseudo-label fitting and network training.

\paragraph{Torso extraction.}
Starting from the SMPL-X template, we keep the vertices and faces corresponding to the upper torso, neck, shoulders, and a short upper-arm boundary. Lower-body and hand regions are removed to reduce unnecessary degrees of freedom for portrait images. The remaining torso keeps the SMPL-X shape blendshapes and the spine/neck/shoulder joints used by the portrait kinematic tree. 
% \exptodo{confirm the exact retained SMPL-X vertex/face index set and whether upper-arm vertices are kept only as boundary support or predicted at test time.}

\paragraph{Canonical alignment.}
For every shape instance, we instantiate the FLAME head and the SMPL-X torso in zero global pose. We compute the FLAME eye-center anchor $C_h$ from the left/right eyeball or eye-socket landmarks and the torso eye-center anchor $C_p$ from the corresponding SMPL-X head landmarks before head removal. The FLAME head is translated by $C_p-C_h$ before injection. This eye-center alignment preserves face scale and keeps the FLAME head in the SMPL-X camera-relative coordinate system. 
% \exptodo{confirm whether a scale correction or rigid Procrustes alignment is used in addition to translation.}

\paragraph{Topology injection.}
The original SMPL-X head faces are removed above the neck connection band, and the aligned FLAME head vertices/faces are inserted. Around the neck transition, we use a fixed connection band $\mathcal{R}$ containing paired FLAME and torso boundary vertices. The final connection vertices are linearly blended from the two templates,
\begin{equation}
    v_i = (1-\alpha_i)v_i^{\mathrm{FLAME}} + \alpha_i v_i^{\mathrm{SMPL\text{-}T}},
    \quad i \in \mathcal{R},
\end{equation}
where $\alpha_i$ increases from the upper neck/head side to the lower neck/torso side. Faces in the transition band are rebuilt from a fixed triangulation so that the output template is watertight and has constant topology across frames. The neck boundary corresponds to FLAME's bottomline of 30 vertices. The default model uses no transition bands; the soft-stitch variant uses one band (neck\_upper) with geometric width 0.015 m. Mesh faces are manually predefined from the template assets; nearest-neighbor matching is used only for blend weights, not for triangulation.

\paragraph{Skinning weights.}
For vertices inherited from SMPL-X, we keep the original SMPL-X linear-blend-skinning weights. For FLAME face vertices, rigid head-region weights are assigned to the head joint except for jaw, eyeball, and transition-band vertices: FLAME jaw vertices keep jaw-dependent deformation, eyeball vertices follow the eye joints, and transition-band weights are interpolated from neighboring head and neck/torso weights. This preserves FLAME expression and jaw motion while making global head motion a child of the SMPL-X neck chain. FLAME jaw/eye skinning weights are copied directly from FLAME2023.

\paragraph{Forward pass.}
At inference, the network predicts $(\boldsymbol{\beta}_h,\boldsymbol{\beta}_t,\boldsymbol{\psi},\boldsymbol{\theta},\boldsymbol{\pi})$. PPM first builds the canonical injected template with FLAME shape/expression/jaw and SMPL-X torso shape, then applies the unified LBS under the spine $\rightarrow$ neck $\rightarrow$ head $\rightarrow$ jaw/eye hierarchy. Finally, vertices are projected with camera $\boldsymbol{\pi}$. Because the injection is performed in canonical space, the neck connection is independent of current head pose and remains stable under animation.

\paragraph{Computation and Manual Refinement of Barycentric Coordinates.}

To enable PPM to support projection of Sapiens2 facial and shoulder keypoints, we adopt a two-stage approach to compute the barycentric coordinates for each keypoint on the PPM mesh. First, we select a set of clear images as candidates. For each image, we use the Sapiens2 pose estimator to predict facial and upper body keypoints, and simultaneously use the data generation pipeline described in the next section (excluding losses related to Sapiens2 keypoints) to obtain a mesh that is aligned to the image. By projecting the estimated keypoints onto the mesh, we obtain coarse barycentric coordinates for each point. 

Next, we develop a refinement tool that enables manual adjustment of these barycentric coordinates, ensuring that they precisely correspond to the Sapiens2 facial and shoulder keypoint positions and are consistent and accurate in 3D.

\begin{figure}[htbp]
    \centering
    \includegraphics[width=\textwidth]{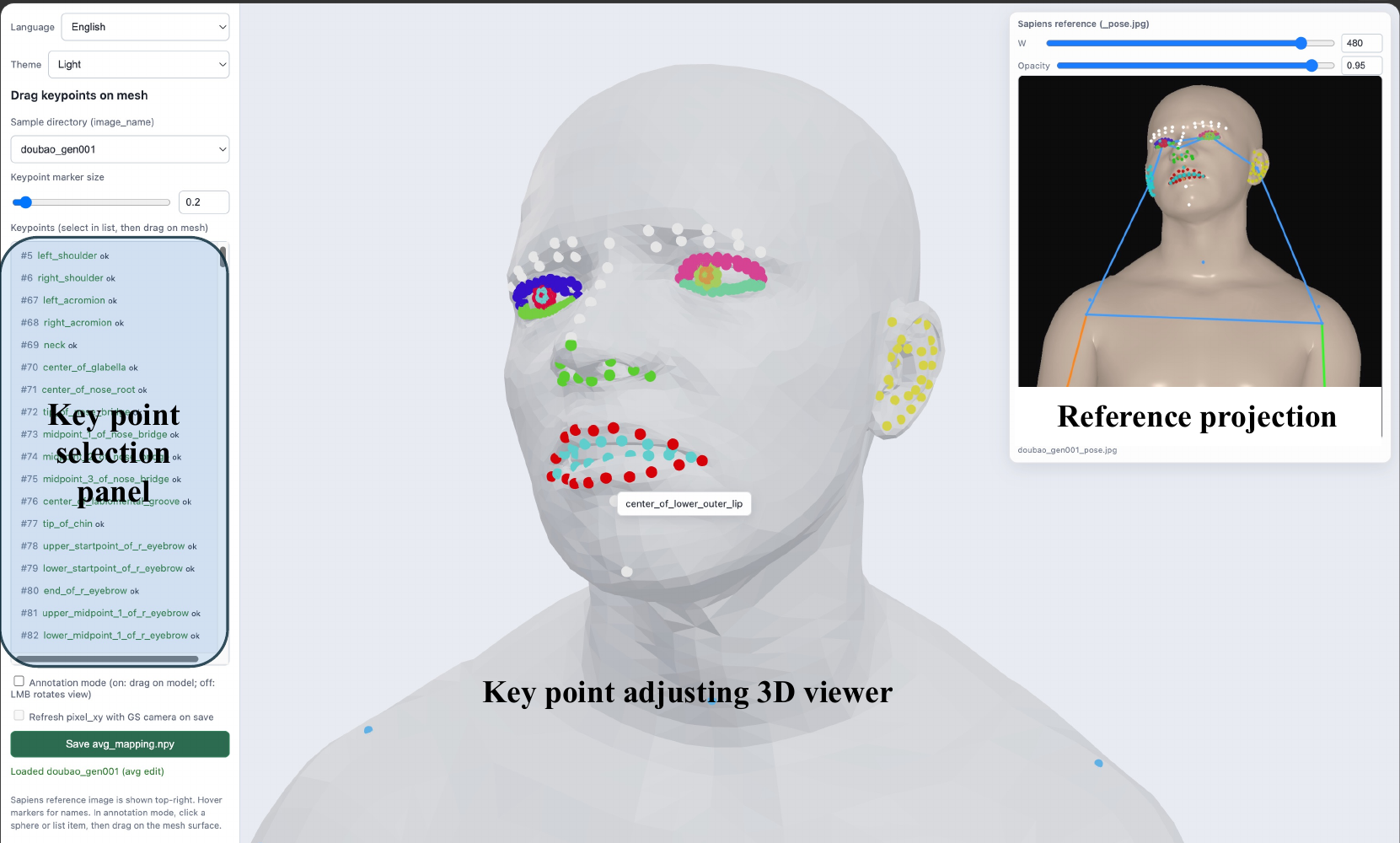}
    \caption{User interface for manually adjusting the barycentric coordinates of Sapiens2 keypoints on the PPM mesh. The tool allows refining the projected positions to ensure precise correspondence and 3D consistency.}
\end{figure}

\section{Data Generation Pipeline}
\label{sec:supp_data_pipeline}

\begin{figure}[htbp]
    \centering
    \includegraphics[width=\linewidth]{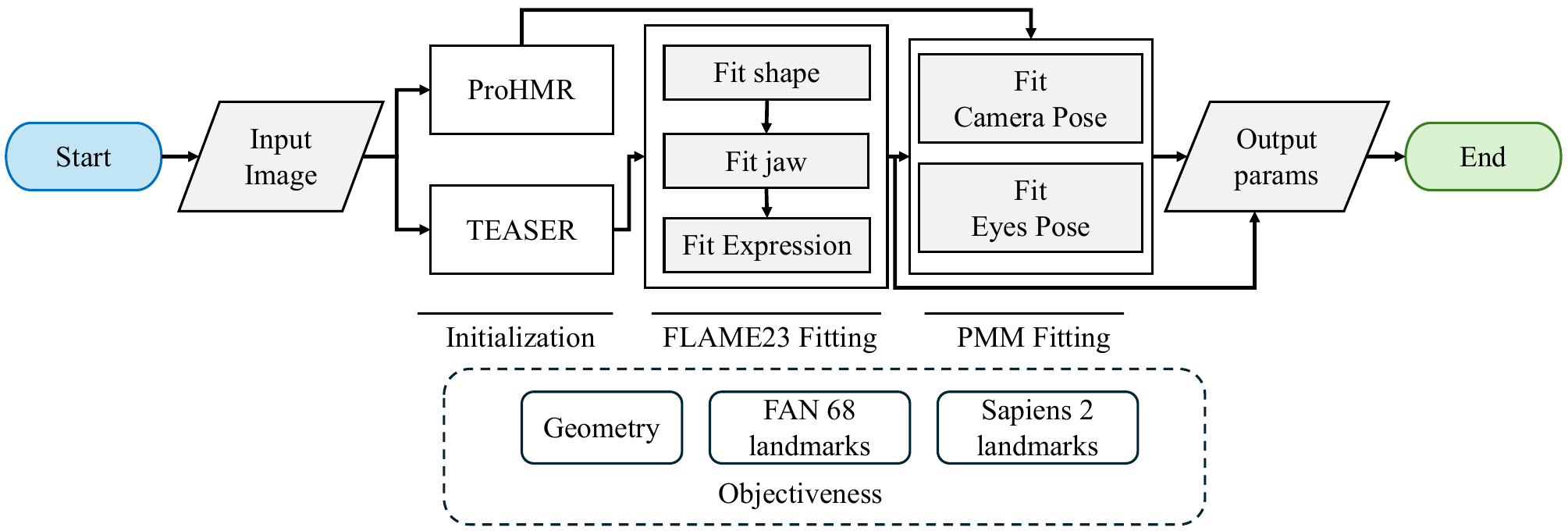}
    \caption{\textbf{Graduated-Refinement Pipeline.} We initialize body and face separately, then fit FLAME2023 in three steps (shape $\rightarrow$ jaw $\rightarrow$ expression), and finally integrate the result into PPM.}
    \label{fig:data_fit}
\end{figure}

Training PAA requires pseudo-labels with decoupled head pose, jaw motion, and expression. Most public datasets do not provide this, and joint fitting often mixes jaw rotation with expression blendshapes. We therefore run an offline \textbf{Graduated-Refinement Pipeline} on in-the-wild images (\cref{fig:data_fit}). The pipeline follows the same coarse-to-fine order as training: fit shape, then jaw, then expression, then full PPM parameters.

Each image $\mathbf{I}$ is processed in three phases: initialization, graduated FLAME fitting, and PPM integration.

\subsection{Phase 1: Hybrid Initialization}
Fitting is sensitive to initialization. We warm-start from two regressors:
\begin{itemize}
    \item \textbf{Body and camera (ProHMR):} ProHMR~\cite{Kolotouros2021ProHMR} gives initial camera $\boldsymbol{\pi}_{\mathrm{init}}$ and torso shape $\boldsymbol{\beta}_{t,\mathrm{init}}$.
    \item \textbf{Face (TEASER):} TEASER~\cite{Liu2025TEASER} gives initial head shape $\boldsymbol{\beta}_{h,\mathrm{init}}$, expression $\boldsymbol{\psi}_{\mathrm{init}}$, and head-related pose.
\end{itemize}

\subsection{Phase 2: Graduated FLAME-2023 Fitting}
We fit FLAME2023 in three separate steps to reduce jaw--expression leakage (\cref{fig:data_fit}, center):
\begin{equation}
    \mathbf{P}_{\mathrm{face}} = \mathrm{FLAME}(\boldsymbol{\beta}_h, \boldsymbol{\psi}, \boldsymbol{\theta}_{\mathrm{jaw}}).
\end{equation}

\noindent\textbf{Step 2.1: Shape.}
Fix pose and expression to their initial values and optimize head shape only:
\begin{equation}
    \min_{\boldsymbol{\beta}_h} \mathcal{L}_{\mathrm{lmk}}\!\left(\mathrm{FLAME}(\boldsymbol{\beta}_h, \boldsymbol{\psi}_{\mathrm{init}}, \boldsymbol{\theta}_{\mathrm{init}}), \mathbf{P}_{2d}\right) + \lambda_{\mathrm{reg}} \|\boldsymbol{\beta}_h\|_2^2.
\end{equation}

\noindent\textbf{Step 2.2: Jaw.}
Fix $\boldsymbol{\beta}_h$ and set mouth-related expression coefficients to zero. Optimize jaw rotation $\boldsymbol{\theta}_{\mathrm{jaw}}$ so mouth opening is explained by the jaw joint rather than expression.

\noindent\textbf{Step 2.3: Expression.}
Fix $\boldsymbol{\beta}_h$ and $\boldsymbol{\theta}_{\mathrm{jaw}}$, then optimize $\boldsymbol{\psi}$ for remaining surface detail (e.g., lip compression and cheek motion).

\subsection{Phase 3: PPM Integration}
We merge the fitted FLAME head into PPM and refine global pose.

\noindent\textbf{Camera and torso.}
Using ProHMR body cues and the fitted head, we optimize $\boldsymbol{\pi}$, $\boldsymbol{\beta}_t$, and torso-related pose components in $\boldsymbol{\theta}$ (spine and neck). This aligns the spine $\rightarrow$ neck $\rightarrow$ head chain with the image.

\noindent\textbf{Eyes.}
We refine eye gaze $\boldsymbol{\theta}_{\mathrm{eye}}$ to match iris landmarks.

The final pseudo-label is
\begin{equation}
    \mathbf{P}_{\mathrm{out}} = \{\boldsymbol{\pi}, \boldsymbol{\beta}_h, \boldsymbol{\beta}_t, \boldsymbol{\theta}, \boldsymbol{\psi}\},
\end{equation}
which matches the parameterization in \cref{subsec:ppm} and is used to supervise the HKD-Exp Network during PAA training.

% !TEX root = ../main.tex
\section{Additional Experimental Results}
\label{sec:exp_supp}

\subsection{Full Component Ablation}
\label{subsec:ablation_full}

We study training objectives and Graduated-Mask Router settings to answer \textbf{RQ4} (\cref{tab:ablation_components}).
Early rows (V0.x) replace the backbone or add data augmentation.
Later rows add losses one by one, then change the router activation $\alpha_{\mathrm{act}}$.
Among the reported JSR and Ner-MVE cells, V2.2 (Our final, $\alpha_{\mathrm{act}}{=}0.50$) reaches JSR $0.799$ and Ner-MVE $8.20$, which matches the full model in the main tables.

% TODO: re-run ablation with current GRAPE recipe (Pixel3DMM FeatDistill, 224 crop).
% Legacy NME values omitted; do not compare with Tab.~\ref{tab:cmp_landmark}.
\begin{table}[t]
  \centering
  \caption{{Ablation on training objectives and Graduated-Mask Router.
    $\mathcal{L}_{\mathrm{ord}}$ / $\mathcal{L}_{\mathrm{fd}}$ / $\mathcal{L}_{\mathrm{nc}}$ / $\mathcal{L}_{\mathrm{om}}$ / $\mathcal{L}_{\mathrm{rg}}$ follow the main loss terms;
    \textbf{Aug}: training-time augmentation;
    $\alpha_{\mathrm{act}}$: Bernoulli activation probability in the Graduated-Mask Router.}}
  \label{tab:ablation_components}
  \small
  \setlength{\tabcolsep}{3pt}
  \begin{tabular}{lccccccccccc}
    \toprule
    Variant
      & Aug & $\mathcal{L}_{ord}$ & $\mathcal{L}_{fd}$ & $\mathcal{L}_{nc}$ & $\mathcal{L}_{om}$ & $\mathcal{L}_{rg}$
      & $\alpha_{\mathrm{act}}$
      & NME-P\ \mdown
      & NME-S\ \mdown
      & \dsHdtf{} JSR \mup
      & Ner-MVE \mdown \\
    \midrule
    % AUTO_START tab_ablation_components
V0.1 (ViT w/o DA) &  &  &  &  &  &  & 1.0 & 0.141 & 0.559 & 0.270 & 37.1 \\
V0.2 (ViT with DA) & \checkmark &  &  &  &  &  & 1.0 & 0.127 & 0.300 & 0.270 & 28.3 \\
V0.3 (Sapiens2 Encoder) & \checkmark &  &  &  &  &  & 1.0 & 0.096 & 0.207 & 0.366 & 21.9 \\
V1.1 & \checkmark & \checkmark &  &  &  &  & 1.0 & 0.062 & 0.176 & 0.566 & 21.0 \\
V1.2 & \checkmark & \checkmark & \checkmark &  &  &  & 1.0 & 0.064 & 0.172 & 0.686 & 14.4 \\
V1.3 & \checkmark & \checkmark & \checkmark & \checkmark &  &  & 1.0 & 0.064 & 0.174 & 0.711 & 15.1 \\
V1.4 & \checkmark & \checkmark & \checkmark & \checkmark & \checkmark &  & 1.0 & 0.060 & 0.165 & 0.708 & 8.40 \\
V1.5 & \checkmark & \checkmark & \checkmark & \checkmark & \checkmark & \checkmark & 1.0 & 0.059 & 0.153 & 0.701 & 8.33 \\
V2.1 & \checkmark & \checkmark & \checkmark & \checkmark & \checkmark & \checkmark & 0.75 & 0.058 & 0.156 & 0.784 & 8.19 \\
V2.2 (Our final) & \checkmark & \checkmark & \checkmark & \checkmark & \checkmark & \checkmark & 0.50 & 0.058 & 0.155 & 0.799 & 8.20 \\
V2.3 & \checkmark & \checkmark & \checkmark & \checkmark & \checkmark & \checkmark & 0.25 & 0.066 & 0.165 & 0.801 & 8.58 \\
    % AUTO_END tab_ablation_components
    \bottomrule
  \end{tabular}
\end{table}

\begin{figure}[htbp]
    \centering
    \includegraphics[width=\linewidth]{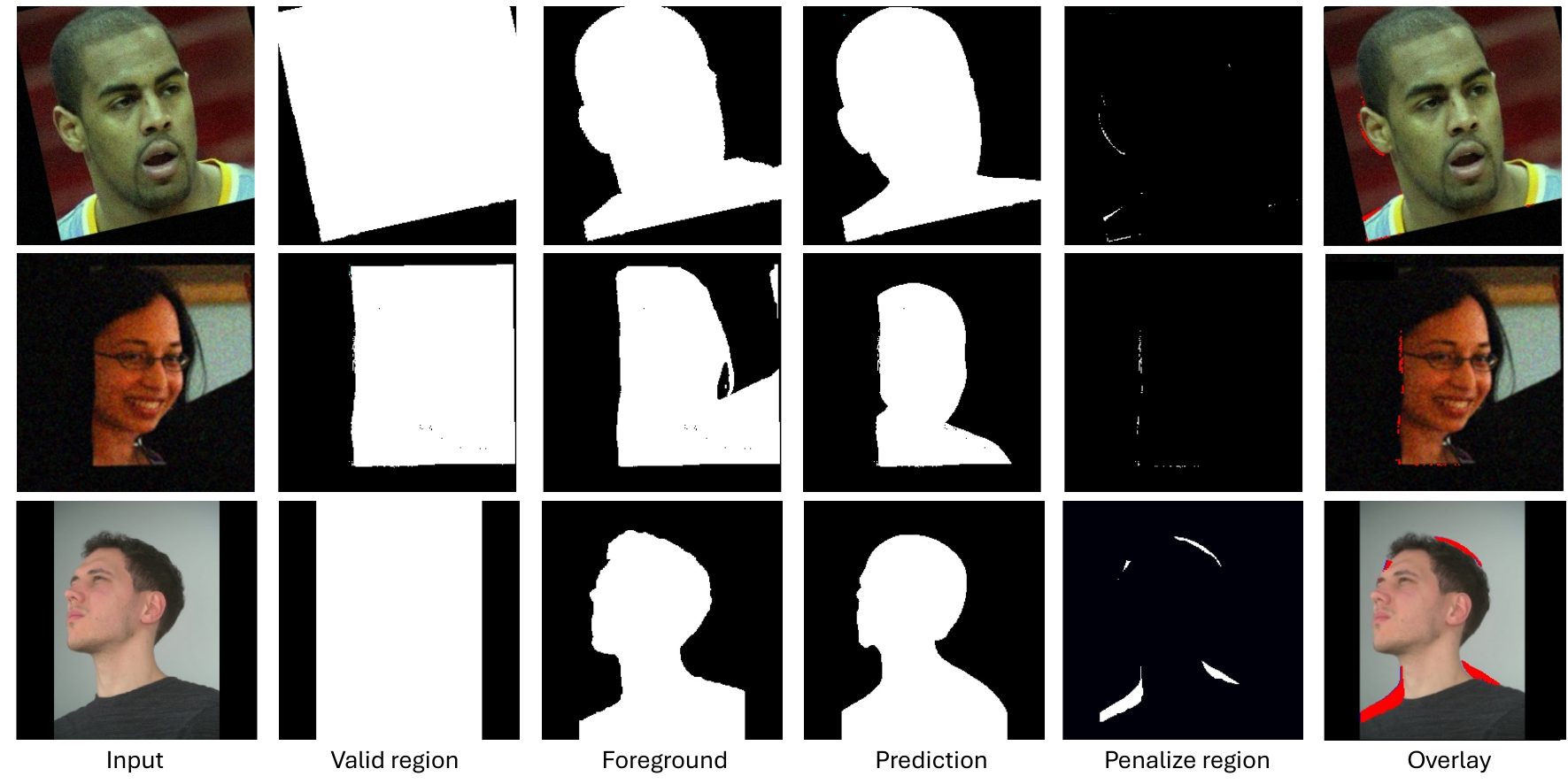}
    \caption{{Visualization of the overflow mask used by $\mathcal{L}_{\mathrm{om}}$.
    The mask marks projected torso regions that fall outside the portrait foreground.}}
    \label{fig:overfull_mask_viz}
\end{figure}

\cref{fig:overfull_mask_viz} shows the overflow mask used in $\mathcal{L}_{\mathrm{om}}$. It penalizes projected torso pixels outside the foreground mask and helps keep the body mesh inside the visible portrait region.

\subsection{Jaw--Skeletal Ratio Analysis}
\label{subsec:analysis_jsc_clip}

\cref{fig:jaw_distance_analysis} (main paper) visualizes jaw--skeletal coupling on a representative talking clip.
For each frame $t$, we compute the \emph{jaw-chain ratio} (JCR) from the projected nose tip $\mathbf{u}^{t}_{\mathrm{nose}}$ and chin tip $\mathbf{u}^{t}_{\mathrm{chin}}$:
\begin{equation}
    \mathrm{JCR}_t =
    \frac{\|\mathbf{u}^{t}_{\mathrm{nose}}-\mathbf{u}^{t}_{\mathrm{chin}}\|_2 - \mu_d}
    {\sigma_d + \epsilon},
    \quad
    d_t=\|\mathbf{u}^{t}_{\mathrm{nose}}-\mathbf{u}^{t}_{\mathrm{chin}}\|_2,
\end{equation}
where $\mu_d$ and $\sigma_d$ are computed over the evaluated speech frames of the same clip. JCR is a normalized nose--chin distance reference. It reduces subject-scale and crop-size effects while keeping the skeletal opening trend from jaw motion. We define the \emph{jaw--skeletal ratio score} (JSR) as
\begin{equation}
    \mathrm{JSR} =
    \rho\!\left(\{\theta^{t}_{j-p}\}_{t=1}^{T}, \{\mathrm{JCR}_t\}_{t=1}^{T}\right),
\end{equation}
where $\theta^{t}_{j-p}$ is predicted jaw pitch and $\rho(\cdot,\cdot)$ is Pearson correlation. Unlike landmark-based mouth-opening correlations, JSR does not reward expression shortcuts that only mimic mouth opening in 2D landmarks. The main paper reports HDTF-wide JSR in \cref{tab:cmp_hdtf}. 

\subsection{Metric Definitions and Evaluation}
\label{subsec:metric_defs}

\paragraph{Metric Definitions.} The metrics are defined as follows:
\begin{itemize}
  \item \textbf{68pt NME} ($\downarrow$): inter-ocular normalized mean error in \%, excluding jawline indices 0--16~\cite{wu2026pear}.
  \item \textbf{Sapiens-Face / Shoulder} ($\downarrow$): mean L2 in normalized crop coordinates on Sapiens keypoint subsets~\cite{sapiens2}.
  \item \textbf{JCR}: normalized nose--chin distance reference computed within each talking clip.
  It is used as a skeletal mouth-opening proxy on clips with relatively stable head pose.
  \item \textbf{JSR} ($\uparrow$): Pearson $r$ between predicted jaw pitch and JCR.
  Higher JSR indicates that jaw articulation follows the skeletal opening trend rather than expression shortcuts (\cref{fig:jaw_distance_analysis}, \cref{subsec:analysis_jsc_clip}).
  \item \textbf{Temporal} ($\downarrow$): mean second-order temporal difference of jaw and expression parameters.
  \item \textbf{NME-P / NME-S} ($\downarrow$): projection landmark errors on face / shoulder regions in the talking-head eval crop.
  \item \textbf{MVE / LVE} ($\downarrow$): mean / lip vertex error on method-specific head mesh topologies~\cite{feng2021learning}.
  \item \textbf{MPJPE} ($\downarrow$): head--shoulder joint error in millimeters~\cite{kolotouros2019learning,Liu2025TEASER}.
\end{itemize}

\paragraph{Evaluation Protocol.}
All image and video baselines use their own official crop protocols for fair comparison (\cref{subsec:setup}). HDTF~\cite{zhang2021flow} evaluation uses $20$ randomly selected videos, consistent with the main paper. Nersemble mesh comparison may mix FLAME2020 (e.g., TEASER~\cite{Liu2025TEASER}) and FLAME23 (\methodname{}~\cite{flame23}).

\subsection{More Downstream Results}
\label{subsec:more_apps}

\begin{figure}[ht]
  \centering
  \includegraphics[width=\linewidth]{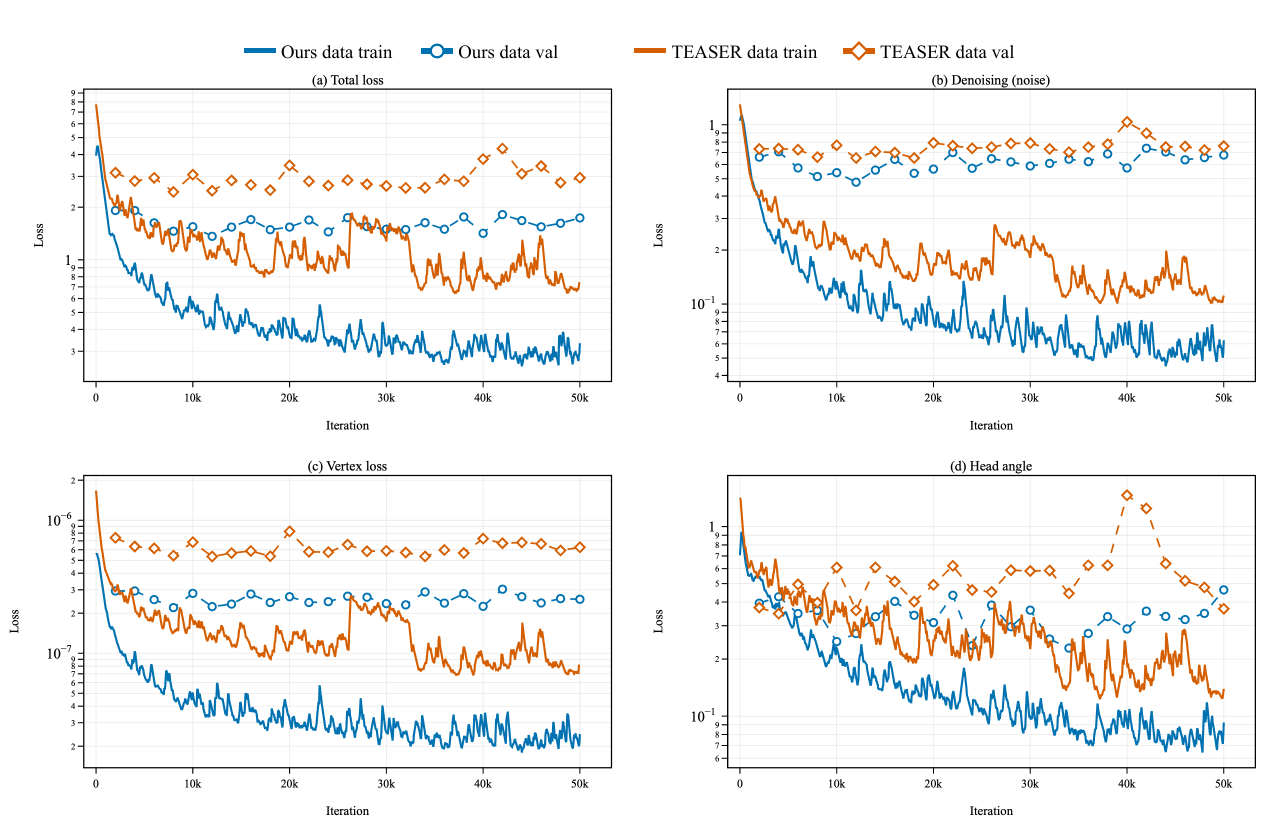}
  \caption{Additional DiffPoseTalk~\cite{sun2024diffposetalk} training curves on \dsHdtf{}.
  We compare TEASER-generated parameters and \methodname{}-generated parameters.}
  \label{fig:supp_dpt_train}
\end{figure}

We provide additional DiffPoseTalk~\cite{sun2024diffposetalk} training curves in \cref{fig:supp_dpt_train}.
As in the main paper (\cref{fig:talking_head}), we train on TEASER-generated parameters and on \methodname{}-generated parameters.
The curves again show more stable training with the \methodname{}-labeled set.

We further report quantitative RGBAvatar~\cite{li2025rgbavatar} results in \cref{tab:cmp_rgbavatar}, corresponding to the qualitative comparison in the main paper (\cref{fig:avatar_rec}).
We randomly select $10$ videos; for each video, the last $350$ frames are used as the test set and the remaining frames as the training set.
Full PPM (\texttt{ppm\_ub}) improves both full-frame and body-region metrics over TEASER and over \methodname{} head-only (\texttt{ppm\_head}). 
The clearest gain is on PSNR-b ($18.64$ vs.\ $17.90$ for TEASER), which is consistent with reduced floating-head artifacts in the neck--shoulder region.

\begin{table}[t]
  \centering
  \caption{Quantitative avatar reconstruction with RGBAvatar~\cite{li2025rgbavatar} on \dsHdtf{}.
    We compare avatars driven by FLAME2020 parameters from TEASER~\cite{Liu2025TEASER},
    \methodname{} head-only parameters (\texttt{ppm\_head}),
    and full upper-body PPM parameters (\texttt{ppm\_ub}).
    \textbf{SSIM} / \textbf{PSNR}: full-frame reconstruction quality ($\uparrow$).
    \textbf{SSIM-b} / \textbf{PSNR-b}: body-region reconstruction quality ($\uparrow$).}
  \label{tab:cmp_rgbavatar}
  \small
  \setlength{\tabcolsep}{8pt}
  \begin{tabular}{lcccc}
    \toprule
    Method & SSIM \mup & PSNR \mup & SSIM-b \mup & PSNR-b \mup \\
    \midrule
    FLAME2020 (TEASER~\cite{Liu2025TEASER}) & 0.844 & 21.08 & 0.673 & 17.90 \\
    PPM head-only (\methodname{}) & \second{0.845} & \second{21.39} & \second{0.687} & \second{18.63} \\
    PPM (\methodname{}) & \best{0.848} & \best{21.69} & \best{0.693} & \best{18.64} \\
    \bottomrule
  \end{tabular}
\end{table}

\section{Limitations, Discussions and Future Work}
\label{sec:discussions}

\paragraph{Limitations.}
\methodname{} targets portrait images where the face, neck, and at least part of the shoulders are visible.
The torso-rooted PPM improves head--neck coherence when these regions are observed.
The benefit can drop when the neck or shoulder is heavily occluded by hair, hands, scarves, or loose clothing.
PPM is a tight-surface parametric model.
Constrained by the FLAME representation, the current PPM does not accurately model teeth, the tongue, or other intra-oral structures, and thus cannot fully capture the full range of portrait expressions.
It does not reconstruct hair volume, garments, jewelry, or other non-body geometry.
In those cases, the projected mesh can still serve as a kinematic scaffold, but it is not a full scene reconstruction.

Training and evaluation also depend on pseudo supervision.
Sapiens-2 keypoints, 68-point landmarks, foreground masks, and offline-fitted PPM labels provide complementary signals.
Each signal can fail under extreme profile views, motion blur, low resolution, unusual lighting, or uncommon poses.
Multi-source supervision reduces reliance on any single cue, but it cannot remove all pseudo-label bias.
JCR/JSR is a weak skeletal proxy for talking clips with relatively stable head pose.
Large head motion, detector jitter, or bad nose/chin localization can affect the score.
JSR should be read together with qualitative results and reconstruction metrics, not alone.

\paragraph{Ethical Considerations.}
Portrait reconstruction and animation can support telepresence, accessibility, digital avatars, and creative tools.
They can also be misused for impersonation or non-consensual manipulation.
Our work focuses on geometric reconstruction and parameter disentanglement, not identity transfer or photorealistic synthesis.
Applications built on \methodname{} should obtain consent, disclose generated or animated content when appropriate, and avoid deceptive or unauthorized biometric uses.
Dataset use should respect licenses and privacy requirements.

\paragraph{Discussions.}
The main design choice is to treat portrait recovery as an articulated portrait problem, not a face-only fitting problem.
PPM closes the representation gap with a torso-rooted reference frame.
Multi-source supervision addresses weak observability of portrait variables.
PAA with the Graduated-Mask Router reduces the factorization gap between rigid articulation and non-rigid expression. These parts work together. A richer model without stronger supervision can still drift in the shoulder region. Stronger supervision without anatomical routing can still fit mouth motion through expression shortcuts. The graduated design is most useful when the output will be animated or edited, where parameter meaning matters beyond per-frame landmark accuracy.

There is also a trade-off between generality and controllability. Implicit or generative representations can capture hair, clothing, and fine appearance more easily, but they often give weaker explicit control over neck, head, and jaw. PPM keeps an explicit kinematic structure for talking-head animation and avatar control, at the cost of modeling only the body-like portrait surface.

\paragraph{Future Work.}
We already show preliminary benefits for DiffPoseTalk and RGBAvatar in the main experiments (\cref{subsec:app}); RGBAvatar quantitative metrics are in \cref{tab:cmp_rgbavatar}. Future work can extend \methodname{} in three directions.
First, combine the PPM scaffold with neural implicit fields or Gaussian splatting to model hair, clothing, and accessories while keeping kinematic control.
Second, improve pseudo-label generation with temporal fitting, multi-view consistency, or uncertainty-aware supervision, so unreliable shoulder and chin cues weigh less during training.
Third, expand disentanglement evaluation beyond JCR/JSR with tracked jaw joints, dental scans, or high-quality 4D face captures.

\end{document}